%% file: main.tex
\documentclass[acmlarge]{acmart}
\usepackage{xcolor}
\usepackage{tikz}
\usepackage{subfig}
\hypersetup{
    colorlinks=true,
    linkcolor=black, % Color for internal links
    citecolor=citeblue, % Color for citation links
    urlcolor=blue % Color for URLs
}
\usepackage{multirow}
\usepackage{tcolorbox}
\usepackage{lipsum}
\usepackage{amsmath, nccmath}
\usepackage{algorithm}
\usepackage{algorithmic}
\usepackage{caption}
\usepackage{lipsum}
\usepackage{refcount}
\usepackage{graphicx}
\usepackage{wrapfig}
\usepackage{breakcites}
\usepackage{pifont}
\definecolor{citeblue}{RGB}{0, 0, 255} % Blue color for citations

\newtcolorbox[auto counter, number within=subsection]{mybox}[2][]{
    title=\thetcbcounter: #2, label=box:#2,#1
}
\newcommand{\patchex}{\textit{PatchEX}}
\newcommand*\circled[1]{\tikz[baseline=(char.base)]{\node[shape=circle,fill,inner sep=0pt,minimum size=1pt] (char) {\textcolor{white}{#1}};}}
\AtBeginDocument{%
  }

\setcopyright{acmlicensed}
\copyrightyear{2024}
\acmYear{2024}
\acmDOI{XXXXXXX.XXXXXXX}

\begin{document}

\title{\patchex: High-Quality Real-Time Temporal Supersampling through Patch-based Parallel Extrapolation}


\author{Akanksha Dixit}
\affiliation{%
  \institution{Electrical Engineering, Indian Institute of Technology Delhi}
  \city{New Delhi}
  \country{India}
}
\email{Akanksha.Dixit@ee.iitd.ac.in}

\author{Smruti R. Sarangi}
\affiliation{%
  \institution{Electrical Engineering, Indian Institute of Technology Delhi}
  \city{New Delhi}
  \country{India}
}
\email{srsarangi@cse.iitd.ac.in}

%\renewcommand{\shortauthors}{ et al.}

% \begin{abstract}
%   Abstract
% \end{abstract}

\input{abstract}





\maketitle

\input{introduction}
\input{relatedwork}
\input{characterization}

\input{overview}

\input{model}
\input{evaluation}
\input{conclusion}


\bibliographystyle{ACM-Reference-Format}
\bibliography{references}

%\appendix

\end{document}

%% file: abstract.tex
\begin{abstract}

High-refresh rate displays have become very popular in recent years due to the need for superior visual quality in gaming,
professional displays and specialized applications like medical imaging. However, high-refresh rate displays alone do
not guarantee a superior visual experience; the GPU needs to render frames at a matching rate.
Otherwise, we observe disconcerting visual artifacts such as
screen tearing and stuttering. Temporal supersampling is an effective technique to increase frame rates by predicting
new frames from other rendered frames. There are two methods in this space: interpolation and
extrapolation. Interpolation-based
methods provide good image quality at the cost of a higher latency because they also require the next
rendered frame. On the other hand, extrapolation methods are much faster at the cost of quality.
This paper introduces \patchex, a novel frame extrapolation method that aims to provide the quality of interpolation
at the speed of extrapolation. It smartly partitions the extrapolation task into
sub-tasks and executes them in parallel to improve both quality and latency. It then uses a patch-based inpainting
method and a custom shadow prediction approach to fuse the generated sub-frames.
This approach significantly reduces the overall latency while maintaining
the quality of the output. Our results demonstrate that \patchex achieves a 65.29\% and 48.46\% improvement in PSNR over the
latest extrapolation methods ExtraNet and ExtraSS, respectively, while being 6$\times$ and 2$\times$ faster, respectively.

\end{abstract}

%% file: introduction.tex
\section{Introduction}
\label{sec:Introduction}
In the last few years, there has been a significant growth in the demand for high-refresh rate displays. Refresh rates
have reached 360 Hz for major monitor brands. Dell's latest monitors can support refresh rates of up to 500
Hz~\cite{2023Alienware}. This surge is driven by the need for enhanced visual quality in
various market segments such as gaming, professional displays 
(used in fields like finance and e-sports) and specialized
applications such as medical imaging and scientific visualization~\cite{gembler2018effects,
murakami2021study,huhti2019effect}. The global gaming monitor market alone was valued at around USD 9.51 billion in 2022
and is projected to grow to approximately USD 16.04 billion by 2030
with a compound annual growth rate (CAGR) of 6.76\%
between 2023 and 2030~\cite{gamingmarket}. The reason for this trend is because low-refresh rate displays may exhibit various visual artifacts
such as judder
(non-continuous motion perception) and motion blur during 
high-speed motion~\cite{han2022assessing}. 
High-frequency displays thus aim to deliver a smooth and seamless experience by eliminating these artifacts. 

\textbf{GPU is the bottleneck:} It is crucial to acknowledge that having high-frequency displays alone may not always guarantee smooth performance unless the frame rendering rate matches the refresh rate.  When the rendering rate is lower than the refresh rate, visual artifacts such as screen tearing and stuttering can occur~\cite{denes2020perceptual}. Therefore, it is essential for the GPU to render frames at a matching rate, which is seldom  feasible.  As graphics engineers continue to incorporate increasingly complex effects into graphics applications to enhance realism, the rendering process becomes more intricate and time consuming.  Several studies have shown the variation in the rendering rate and its impact on the quality of experience for users~\cite{liu2023effects, xu2024user, sabet2020latency}. This necessitates the exploration of strategies to upsample the rendering rate in real-time even for the latest consumer-grade GPUs.

\textbf{Temporal supersampling fills in frames missed by the GPU:} One of the most impactful approaches to increase the
frame rate is {\em temporal supersampling}, which involves predicting frames using information from the next and
previously rendered frames~\cite{extranet,extrass,softmax, jointss, EMA, wu2023adaptive}. The core concept here is that
since rendering new frames is time-consuming, we can expedite the process by predicting new frames from previously
rendered ones or the next frame (in temporal sequence) and interleave the frames at the display device. 
This boosts the frame rate and achieves rate matching.
For temporal supersampling to be effective, it is important to ensure that the prediction latency is shorter
than the rendering time and that the predicted frame is of acceptable quality. Particularly in real-time systems like
virtual reality applications and games, minimizing latency and ensuring good quality are of utmost importance. 

\begin{figure}[!htb]
	\centering
	\includegraphics[width=0.99\columnwidth]{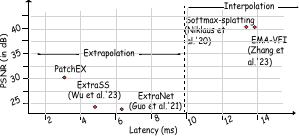}
	\caption{The solution space for temporal supersampling. Each solution is run on an NVIDIA RTX 4090 GPU. The detailed system configuration is shown in Table~\ref{tab:config}.}
	\label{fig:sol_space}
\end{figure}

\textbf{Interpolation: high quality, high latency $\mid$ Extrapolation: low quality, low latency:} 
In the field of temporal supersampling, two primary methodologies exist: \textit{interpolation}~\cite{EMA,softmax,
wu2023adaptive} and \textit{extrapolation}~\cite{extranet,extrass,jointss}. As their names imply,  interpolation
predicts a frame using both past and future frames, whereas extrapolation creates a new frame by utilizing only the past
few frames. Fig.~\ref{fig:sol_space} shows the performance of a few recent works in terms of quality and latency.  It is
evident that interpolation yields superior quality but comes with a higher latency (\textbf{almost 14 ms}, not suitable
for a 90 Hz display), whereas
extrapolation offers lower latency at the expense of inferior quality. This is because interpolation takes into account
both past and future frames (see Section~\ref{subsubsec:inter_vs_extra} for more details). 

Therefore, the challenge is quite
clear: 

\begin{tcolorbox}
  \textbf{Match the quality of interpolation with the performance of extrapolation.}
\end{tcolorbox}

Note that interpolation introduces an inherent latency by holding an already rendered frame for a refresh interval before
displaying it. This is something that the human visual system can easily detect. We tested an interpolation algorithm
on a 360-Hz display and ourselves did not find the quality of the images very appealing.
We thus propose an extrapolation-based approach that does not incur this overhead.
Given that historically such algorithms produced low-quality outputs, real-time extrapolation is a less explored area.
As per our
knowledge, there are only three major works that specifically address this: 
ExtraNet~\cite{extranet}, ExtraSS~\cite{extrass}
and STSS~\cite{jointss}. These methods use a warping algorithm\cite{} that transforms the frame using a motion vector.
Warping algorithms often lead to invalid pixels and holes in certain regions and incorrect shading in other regions. Various
approaches
such as using neural networks  have been employed to rectify these issues. They use the information
stored in G-buffers -- these are data structures in the rendering engine that store
different properties of a scene such as the scene depth, roughness, etc.. Despite
these efforts, none of the methods have produced satisfactory results in complex dynamic environments with multiple
characters and lighting sources. Even though baseline extrapolation methods are fast, the moment neural networks
are added, they become very slow.  

\textbf{Partition and parallelize:} In this paper, we introduce \patchex, a novel approach that is significantly
different from prior frame extrapolation methods. Our methodology involves smartly partitioning the extrapolation task into
sub-tasks and executing them in parallel. The key idea is that we take the warped frame as an initial prediction for
extrapolation. Subsequently, we divide the frame extrapolation task into two primary sub-tasks: fixing invalid pixels in
the warped frame (inpainting) and ensuring proper shadowing in the warped frame. \patchex proposes independent and
simultaneous handling of shadows and inpainting tasks. To achieve this, we remove the shadow information from the frame
and store it in a separate buffer.

The first subtask uses a patch-based inpainting method~\cite{xu2021texture,demir2018patch} to fix the warped frame.
Unlike previous inpainting approaches, we segment the entire frame (without shadows) into three distinct patches/parts
after  taking into account the idiosyncrasies of the human vision system (known as \textit{foveated
segmentation})~\cite{bjorkman2005foveated}. Each patch is subsequently processed in parallel by separate neural
networks that are small in size.
This approach bears a resemblance to classical foveated rendering~\cite{meng2020eye,fov2012}, which aims to
optimize rendering efficiency without compromising visual fidelity. This strategic parallelism drastically reduces the
overall latency. Furthermore, feeding smaller patches to the inpainting networks reduces the inference time.

To handle shadows (second subtask), we create a custom blueprint class in the Unreal Engine (UE)~\cite{2023Unreal} to
extract the shadow map/mask. Using this mask, we predict shadows for the inpainted frame. After the image is inpainted
and shadows are predicted, we merge them to produce the final extrapolated frame.

Currently, we lack large-scale publicly available datasets or workloads for characterizing the real-time rendering
of graphics applications. To address this, we create a dataset by downloading model
and scene files from {\em Epic Games}~\cite{Marketplace} and rendering them using {\em Unreal Engine}
(v5.1)~\cite{2023Unreal}. Our dataset includes multiple animation sequences featuring a diversity of characters,
lighting effects, background scenes and camera motions. 

To summarize, our primary contributions are as follows:\\
\noindent
\circled{1} The first innovation is that we generate a warped frame very quickly to guide the process of accurate frame generation.
\\
\circled{2} Henceforth, the first subtask partitions the warped frame (devoid of shadows) into three different partitions based on the \textit{perceived distance} from the eye. These are rendered differently using different kinds of neural networks. \\
\circled{3} The inpainting method proposed in step (2) runs in parallel for the three kinds of patches (partitions).\\
\circled{4} We meticulously curated a comprehensive dataset featuring a wide range of animation sequences encompassing diverse characters, backgrounds, lighting settings and camera movements.\\
\circled{5} \patchex shows an improvement of 65.29\% and 48.46\% in the PSNR (peak signal-to-noise ratio) compared to the two most recent extrapolation methods, \textit{ExtraNet} and \textit{ExtraSS}, respectively. \\
\circled{6} The proposed inpainting network is 6$\times$ and 2$\times$ faster than the nearest competing works \textit{ExtraNet} and \textit{ExtraSS}, respectively. 

The paper is organized as follows. Section~\ref{sec:RelatedWork} describes the background and related work in the area
of temporal supersampling.  Section~\ref{sec:Characterization} characterizes the datasets and provides the motivation
for the proposed approach. Section~\ref{sec:methodology} presents the methodology in detail. The implementation details
are provided in Section~\ref{sec:Implementation}. Section~\ref{sec:Evaluation} shows the experimental results. We finally
conclude in Section~\ref{sec:Conclusion}.

%% file: relatedwork.tex
\section{Background and Related Work}
\label{sec:RelatedWork}

\subsection{QoE Requirements of Graphics Applications}
\label{subsec:qoe}
Graphics applications need to provide a certain Quality of Experience (QoE) to ensure that users get the best possible
experience. Some of the key QoE requirements are: \circled{1} \textbf{Low latency or high frame rate:} This is important
to ensure that the user sees the effect on the screen as soon as possible after providing an input --
this creates a 
responsive environment that immerses the user in the virtual world. \circled{2} \textbf{High visual quality:} The
rendered frames should be free of visual artifacts such as ghosting, stuttering, motion blurring and screen tearing.  \circled{3} \textbf{Realism:} The frames should
have advanced lighting effects, shadows, and reflections such that the virtual environment appears realistic.
This work aims to
provide all these properties with the available hardware resources of a consumer-grade GPU.

\subsection{Temporal Supersampling in Frame Rendering}
\label{subsubsec:inter_vs_extra}

Recent works primarily focus on \circled{1} predicting new frames using
interpolation~\cite{nehab2007accelerating,herzog2010spatio, andreev2010real} and \circled{2} 
generating new frames using
extrapolation~\cite{extranet} to increase the frame rate. We present a brief comparison of related work in
Table~\ref{table:relwork}. 

\begin{figure}[!htbp]
	\centering
	\includegraphics[width=0.99\columnwidth]{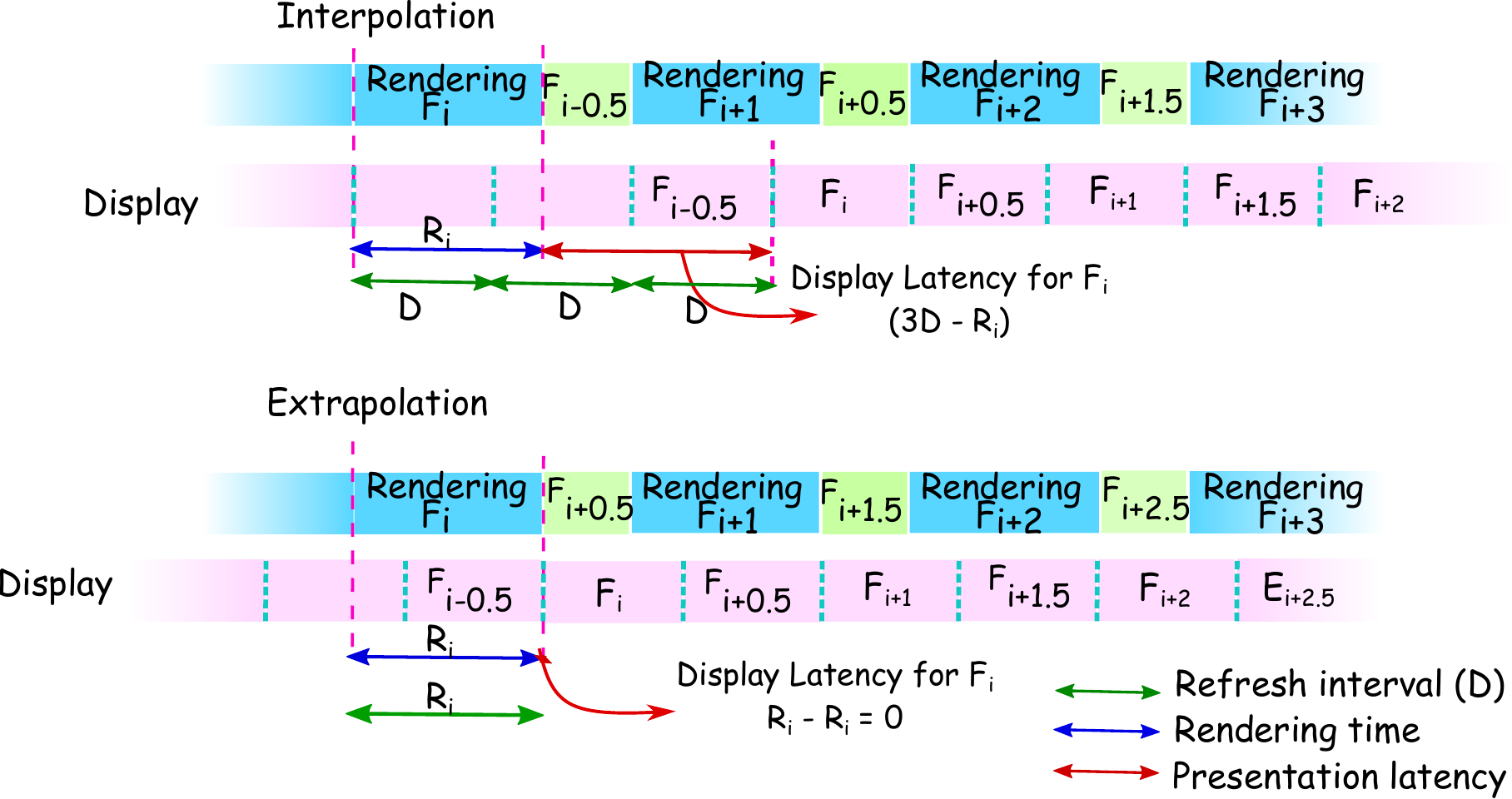}
	\caption{Interpolation and extrapolation explained. 
$F_i$ is the rendered frame. $R$ and $D$ represent the rendering time and refresh latency, respectively.}
	\label{fig:inter_vs_extra}
\end{figure}

As mentioned in Section~\ref{sec:Introduction}, apart from the latency of the algorithm used for interpolation, there is
an additional latency incurred here because \textit{interpolation} predicts frames between two already
\underline{rendered} neighboring frames. We thus need to wait more. In contrast, extrapolation-based methods predict
frames solely based on past frames. The difference can be observed in Fig.~\ref{fig:inter_vs_extra}. Both processes
double the frame rate by generating a new frame after each rendered frame. However, interpolation increases the display
or presentation latency.
In the figure, the presentation latency is the delay between the completion of a frame's rendering
and its actual display on the
screen.  

The mathematical formulae for the presentation latency
of interpolation and extrapolation, respectively, are shown in Table~\ref{tab:lat} (keep referring to Fig.~\ref{fig:inter_vs_extra}). The first assumption is
that the rendering time for every frame is greater than one refresh interval $D$ (in the super-sampled case). If this is not the case, than there
is no need to interpolate or extrapolate in the first place. It is further assumed that
the rendering duration plus the interpolation/extrapolation time does not exceed two refresh intervals $2D$. 
We observe in Fig.~\ref{fig:inter_vs_extra} that if the sum exceeds $2D$, then the interpolated frame will simply not be ready by the time that it needs to be displayed. The assumption here is that we are supersampling
by a factor of $2\times$. We will have similar formulae for other super-sampling ratios.
Our
algorithm per se is not constrained by this choice. The choice of 2 in this example is for the purpose of better explanation.

\begin{table}
\caption{Presentation latency for interpolation and extrapolation}
\begin{center}
\begin{tabular}{|l|l|}
	\hline
	$\forall i, R_i > D$ & assumption \\
	\hline
		$P_i = 3D - R_i$  & \multirow{2}{*}{interpolation}\\
		$R_i + I \le 2D$  &  \\
	\hline
		$P_i = 0$   & \multirow{2}{*}{extrapolation} \\
		$R_{i+1} + E \le 2D$  & \\
	\hline 
\end{tabular}
\end{center}
\label{tab:lat}
\end{table}

$P_i$ and $R_i$ denote the presentation latency and rendering latency for the $i^{th}$ frame, respectively. $D$
represents the refresh interval. $E$ is the latency associated with generating an extrapolated frame (the $(i+0.5)_{th}$
frame) based on frame $F_i$, while $I$ is the latency for generating an interpolated frame using frames  $F_i$ and
$F_{i+1}$.  If we consider a 90 Hz display, the refresh interval $D$ is 11.11 ms. Hence, the presentation latency
$P_i$ for interpolation falls within the range of 11.11 ms to 22.22 ms, which is considerably larger than the latency
for extrapolation, which is 0 (in a system without slack). 

This latency introduced by interpolation significantly affects the user experience due to the human visual system's acute sensitivity to delays. The concept of the just noticeable delay (JND)~\cite{jerald-thesis} underscores this sensitivity, indicating that humans can normally detect delays as low as 3-5 ms with the threshold for gamers and active young people being even lower. It needs to be less than 1 ms in the case of Head-Mounted Displays
(HMDs)~\cite{3ms,haptics}. \textbf{Given these factors, the latency introduced by interpolation (11.11 ms to 22.22 ms) can easily exceed the JND thresholds, leading to perceptible delays and compromising the quality of the viewing experience.}

\begin{table*}[!htb]
	\caption{A comparison of related work}
		\footnotesize
		\begin{center}
	%\resizebox{0.99\textwidth}{!}{
			\begin{tabular}{|l|l|l|l|l|l|l|l|} 
				\hline
				\textbf{Year} &
				\textbf{Work} &
				\textbf{Coherence} &
				\textbf{Method} &
				\textbf{ML-} &
				\textbf{Real-} &
				\textbf{Upsampling} &
				\textbf{Upsampling}\\
				&
				&
				\textbf{Exploited} &
				\textbf{Used} &
				\textbf{based} &
				\textbf{time} &
				\textbf{Domain} &
				\textbf{Factor}\\
				\hline
				2007 & Nehab et al.~\cite{nehab2007accelerating} & Spatio-temporal  & Interpolation & \textcolor{red}{$\times$} & \textcolor{red}{$\times$} & Temporal  & \\
				\hline
				2010 & Andreev et al.~\cite{andreev2010real} & Temporal  & Interpolation & \textcolor{red}{$\times$} & \textcolor{red}{$\times$} & Temporal  & x to 60 \\
				\hline
				2010 & Herzog et al.~\cite{herzog2010spatio} & Spatio-temporal  & Interpolation & \textcolor{red}{$\times$} & \textcolor{red}{$\times$} & Temporal  & \\
				\hline
				2012 & Bowles et al.~\cite{bowles2012iterative} & Temporal  & Interpolation & \textcolor{red}{$\times$}  & \textcolor{red}{$\times$} & Temporal  & \\
				\hline
				2018 & SAS~\cite{mueller2018shading} & Temporal  & Interpolation & \textcolor{red}{$\times$} & \textcolor{red}{$\times$} & Temporal & x to 120 \\
				\hline
				2020 & Softmax-splatting~\cite{softmax} & Temporal  & Interpolation & \textcolor{green}{\checkmark} & \textcolor{red}{$\times$} & Temporal & up to 2 \\
				\hline
				2021 & ExtraNet~\cite{extranet} & Temporal  & Extrapolation & \textcolor{green}{\checkmark} & \textcolor{green}{\checkmark} & Temporal  & up to 2$\times$ \\
				\hline
				2022 &  DLSS 3~\cite{DLSS3} & Spatio-temporal   & Interpolation & \textcolor{green}{\checkmark}  & \textcolor{green}{\checkmark} & Temporal  & up to 4$\times$\\
				\hline
				2023 &  EMA-VFI~\cite{EMA} & Temporal   & Interpolation & \textcolor{green}{\checkmark} & \textcolor{red}{$\times$}  & Temporal  & up to 4$\times$\\
				\hline
				2023 &  ExtraSS~\cite{extrass} & Temporal   & Extrapolation & \textcolor{green}{\checkmark} & \textcolor{green}{\checkmark}  & Spatio-temporal & up to 2$\times$\\
				\hline
				2024 &  STSS~\cite{jointss} & Temporal & Extrapolation & \textcolor{green}{\checkmark} & \textcolor{green}{\checkmark} & Spatio-temporal  & 2$\times$\\
				\hline
				\textbf{2024} & \textbf{Ours} & \textbf{Temporal}  & \textbf{Extrapolation} & \textbf{\textcolor{green}{\checkmark}} &  \textcolor{green}{\checkmark} & \textbf{Temporal} & \textbf{2$\times$} \\
				\hline
			\end{tabular}
	%}
		\end{center}
		
		\label{table:relwork}
\end{table*}
	
\subsubsection{Interpolation} 
\label{subsubsec:interpolation}
Early temporal supersampling-based methods~\cite{nehab2007accelerating} use optical flow-guided interpolation, but
they produce subpar results when the scene contains areas visible in the current frame but not in the previous one.
Although Bowles et al.~\cite{bowles2012iterative} proposed to fix this using an iterative method called fixed point
iteration (FPI), this did not provide satisfactory results. To handle this case, various works ~\cite{DLSS3,
mueller2018shading} propose a bidirectional reprojection method that temporally upsamples frames by reusing data from
both the backward and forward temporal directions. For example, NVIDIA's latest DLSS3 engine does this using
an optical flow generator, a
frame generator and a supersampling network that is AI-accelerated and integrated into its latest GPU architecture
(Ada Lovelace~\cite{DLSS3}). This approach increases the frame rate but also leads to an increased input latency that
users can easily perceive (verified in the lab and reported in the literature\cite{}). 
Since our approach is not based on optical flow fields, it does not require future frames to
predict a new frame. Other methods, such as caching techniques~\cite{nehab2007accelerating} and dividing frames into
slow-moving and fast-moving parts and rendering each part at a different rate~\cite{andreev2010real} have also been
proposed but they increase the time needed to construct a frame significantly. 
Recently, DNN-based solutions~\cite{softmax, EMA} have also been
proposed to produce a high-quality interpolated frame. However, this increases the latency of the interpolation
process due to the complex structure of neural networks. 

\subsubsection{Extrapolation}
\label{subsubsec:extrapolation} 
This is a very sparse area of research. The prominent works that we are aware of are ExtraNet~\cite{extranet},
ExtraSS~\cite{extrass} and STSS~\cite{jointss}.  To predict a new frame, all of these works first propose a warping
algorithm that helps generate a warped frame that serves as an initial prediction. However, the warped frame may have
invalid pixels or holes in disoccluded regions where the temporal information is not available; this leads to incorrect
shading (shadows and reflections) in other regions. To fix the holes, they use a neural network similar to an image
inpainting network~\cite{bertalmio2000image, guillemot2013image}. However, it may not be sufficient in cases where the
shading information changes dynamically over time. To handle this, ExtraNet~\cite{extranet} uses a history encoder to
learn the shading pattern from the previous few frames and fix that in the warped frame. This approach works well when
the lighting conditions change slowly over time. 

On the other hand, STSS~\cite{jointss} uses light source information along with a history encoder. Whereas,
ExtraSS~\cite{extrass} introduces a new warping method to minimize the presence of invalid pixels in the warped frame.
Their technique utilizes G-buffers' information for the warping process. The G-buffer is a set of render targets that
store various properties of a scene during the rendering process. It then uses a
lightweight neural network to fine-tune the shading. Sadly, none of these works produce a satisfactory result in a
complex dynamic environment with numerous characters and lighting sources. Apart from the quality issue, the latency of
these methods due to their heavy neural networks
is significant. Note that the latest work ExtraNet~\cite{extranet} is a pure
temporal supersampling approach, while the remaining two~\cite{jointss, extrass} propose a joint neural network for
supersampling in both temporal and spatial domains. We focus our research solely on the temporal domain. We can 
use a complementary spatial supersampling method if there is a need to increase the resolution of our generated images. 

\subsection{Image Warping}
\label{subsec:warping}

Image warping is a technique that changes the shape or appearance of an image by applying a spatial transformation. This
transformation can include rotation, scaling, translation or other complex deformations. Recent works use motion
vector-based warping~\cite{extranet,extrass,jointss}, where motion vectors encode the displacement of blocks or
regions across frames. These vectors consist of horizontal and vertical components that indicate the amount of movement
in the $x$ and $y$ directions, respectively. There are two types of warping techniques: forward warping and backward
warping~\cite{shimizu2022forward,lee2018iterative}.

In forward warping, each pixel in the frame that needs to be warped (the source frame) is directly mapped to a
corresponding position in the warped frame using the motion information. On the other hand, backward warping involves traversing
each pixel in the target frame and finding its corresponding position in the source frame by applying the inverse
transformation~\cite{zhang2003accelerated}. Each algorithm has its pros and cons. Forward warping is simpler than backward warping since it directly uses the transformation function to map source pixels to the target. However, this direct transformation may result in overlaps because multiple source pixels may mapped to the same target pixel. Additionally, forward warping can create empty spaces or holes in the target frame if all the target pixels do not
receive a mapped value from the source image. In backward warping, although it is possible that a target pixel might not find a corresponding source pixel, resulting in holes, these holes are filled using the values of the nearest source pixels.
Therefore, backward warping produces better quality images
as compared to forward warping because it ensures that every pixel in the target image is assigned a value, avoiding gaps or holes. Forward warping, on the other hand, handles gaps through post-processing interpolation after the warping is completed~\cite{extranet,lee2018iterative}.

Using motion vectors results in residual frames or trails of moving objects in the warped frame,
also known as the {\em ghosting} effect. This occurs because motion vectors store pixel displacements between
consecutive frames but fail to provide information about disoccluded regions. Traditional algorithms in this space
propose to use the
same pixel values from previous frames in these regions, resulting in ghosting. To address this issue, Zeng et
al.~\cite{zeng2021temporally} have proposed a method for generating occlusion motion vectors. These vectors calculate
displacements in disoccluded regions as displacements of nearby regions in the previous frame. But it still fails when
the background becomes complex. To overcome this challenge, Wu et al.~\cite{extrass} introduced a novel technique called
G-buffer-guided warping. This method utilizes a joint bilateral filter that computes the warped pixels, enabling more
accurate and reliable tracking of complex movements. Specifically, it considers a large area of pixels near the warped
pixel and uses weighted G-buffers’ values to blend them to form the warped pixel. As a result, G-buffer-guided warping
has emerged as an effective warping method. Hence, we adopt this technique to generate inputs for our
inpainting network.

\subsubsection{Structure of a G-Buffer}
\label{subsubsec:struc_gbuf}
In the Unreal Engine, a G-buffer (Geometry Buffer) is a set of render targets that store various pieces of information about
the geometry such as the world normal, base color, roughness, textures, etc. During the lighting calculation phase, the
Unreal Engine samples these buffers to determine the final shading of the scene. The G-buffer generally consists of
several textures in the standard RGBA format. However,  Unreal optimizes the performance by packing these attributes
into fewer textures by combining different channels. The exact composition of the G-buffer can vary with the number of
channels. A common example is a five-texture \textit{GBuffer}, which consists of five buffers: $A$ through $E$
(refer to Table~\ref{tab:gbuffer}). Specifically, \textit{GBufferA.rgb} stores the base color with
\textit{transparency} filling the alpha channel. \textit{GBufferB.rgba} stores the 
properties -- metallic, specular and roughness -- and the scene
depth. \textit{GBufferC.rgb} stores the world normal (WN) vector with \textit{GBufferAO} (AO: Ambient Occlusion) filling
the alpha channel. \textit{GBufferD} is dedicated to custom data (stencil buffer) and \textit{GBufferE} is for
precomputed shadow factors.

\begin{table}[!htbp]
	\centering
	\footnotesize
	\resizebox{0.99\columnwidth}{!}{
	\begin{tabular}{|c|c|c|c|c|}
	\hline
	\textbf{G-Buffer} & \textbf{R} & \textbf{G} & \textbf{B} & \textbf{A} \\
	\hline
	\textbf{A} & Base Color (R) & Base Color (G) & Base Color (B) & Transparency \\
	\hline
	\textbf{B} & Metallic & Specular & Roughness & Depth \\
	\hline
	\textbf{C} & WN (X) & WN (Y) & WN (Z) & AO \\
	\hline
	\textbf{D} & \multicolumn{4}{c|}{Custom data (stencil buffer)} \\
	\hline
	\textbf{E} & \multicolumn{4}{c|}{Precomputed shadow factors} \\
	\hline
	\end{tabular}
	}
	\caption{G-buffer layout in Unreal}
	\label{tab:gbuffer}
	\end{table}

\subsection{Foreground Bias Effect in Human Vision}
\label{subsec:foreground_bias}

The human visual system is an incredibly complex and sophisticated mechanism responsible for perceiving and interpreting
visual information.  The process of generating new frames can be challenging due to the intricate nature of this system,
which is highly sensitive to even the slightest input latency and may experience
jitter~\cite{ng2012designing,weier2017perception}. However, certain characteristics of the human vision system can be
leveraged to optimize the frame generation process. One of these characteristics is known as ``\textit{foreground
bias}'' or ``\textit{foreground dominance}''~\cite{fernandes2021foreground}. This phenomenon occurs because the human
visual system tends to focus more on objects in the foreground than those in the background. The primary cause is that
the foreground objects are usually closer to the observer than background elements, this leads to a greater disparity in
the retinal image size. It also provides stronger depth cues (refer to Fig.~\ref{fig:be}). Our visual system is highly
sensitive to depth cues, which contribute to the perceptual salience of foreground objects compared to background
environments. The same effect in game engines like Unity and Unreal Engine are emulated using a technique known as
\textit{Perspective Projection}~\cite{toth2016comparison} making far-away objects appear smaller and fainter than foreground
objects.

The foreground bias effect can be used to our advantage by extrapolating foreground interactions more efficiently since
artifacts are more noticeable in the foreground as compared to the background. 

%% file: characterization.tex
\section{Characterization and Motivation}
\label{sec:Characterization}
In this section, we begin by presenting the benchmarks used in our experiments. Next, we evaluate the frame rendering
times and pinpoint the factors contributing to the latency overhead and frame rate variability. We then demonstrate how
any frame can be segmented into three distinct regions with varying levels of detail, aligning with the viewer's
perception of changes within each region. This {\em strategic segmentation forms a pivotal element of our approach}. Finally,
we address the challenges inherent in the extrapolation process.

\subsection{Overview of the Datasets}
\label{subsec:data}
We render our datasets using Unreal Engine 5 (UE5 v5.1) on an NVIDIA RTX series GPU with the Ada Lovelace architecture.
The detailed configuration is shown in Table~\ref{tab:config}. To create our animation sequences, we downloaded Unreal
scene files from the UE Marketplace~\cite{Marketplace}. We further complicated it by integrating
animations with characters from Mixamo~\cite{mixamo}
into the background scenes to generate various animation sequences.

\begin{figure*}[!h]
	  \centering
	  \subfloat[Bunker]{
		\includegraphics[width=0.245\textwidth]{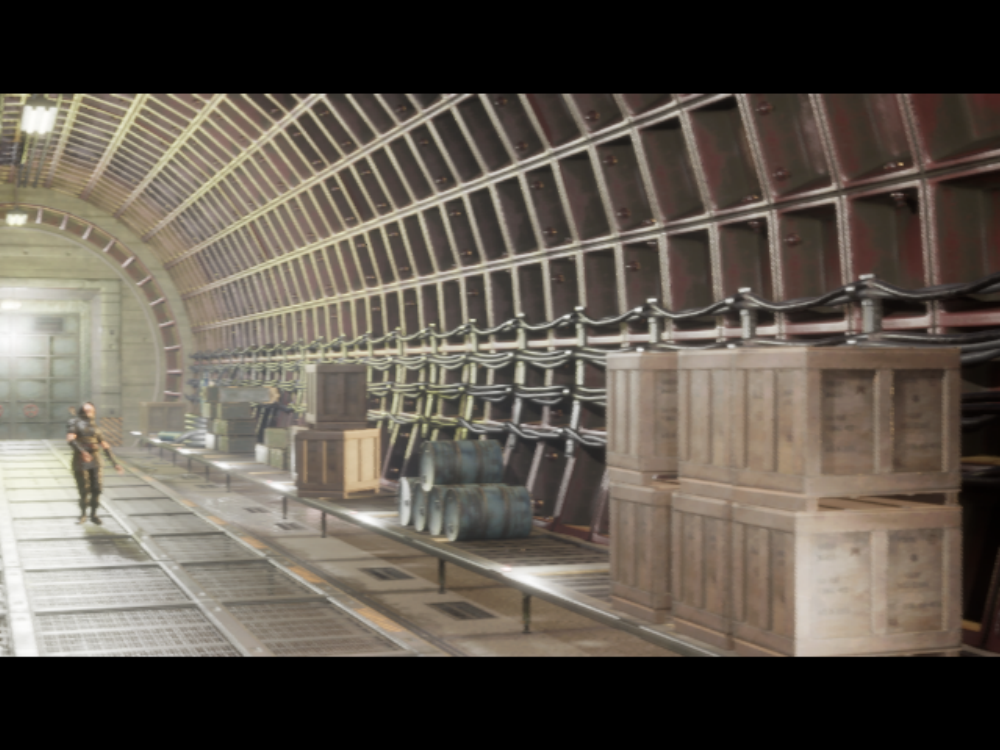}
	  }
	  \subfloat[City Park]{
		\includegraphics[width=0.245\textwidth]{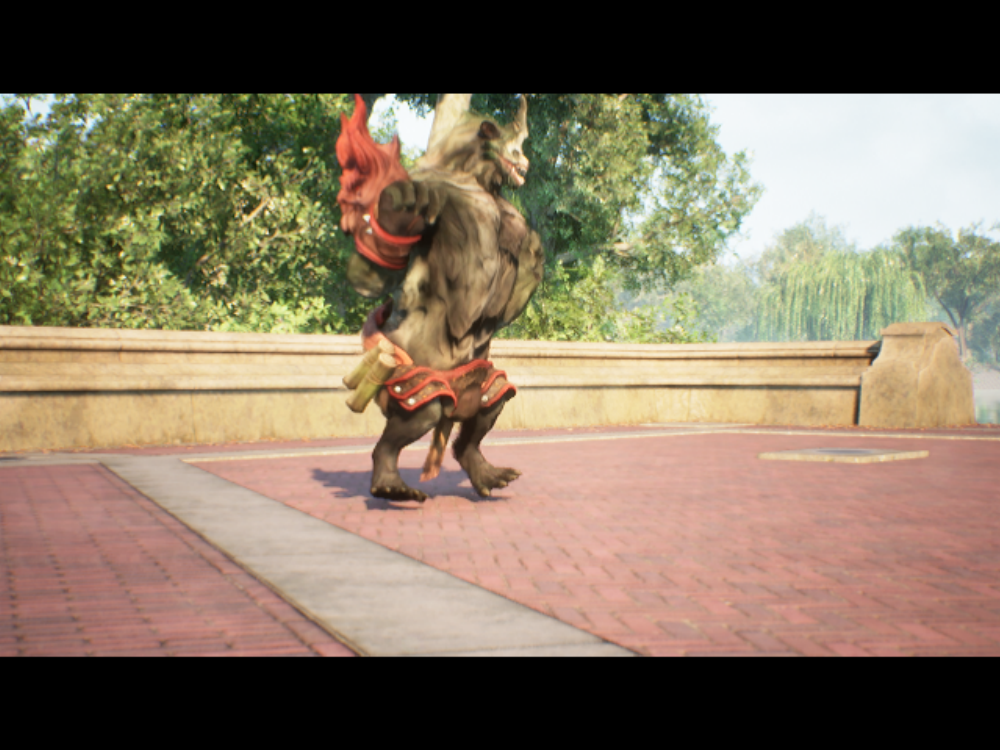}
	  }
	  \subfloat[Western Town]{
		\includegraphics[width=0.245\textwidth]{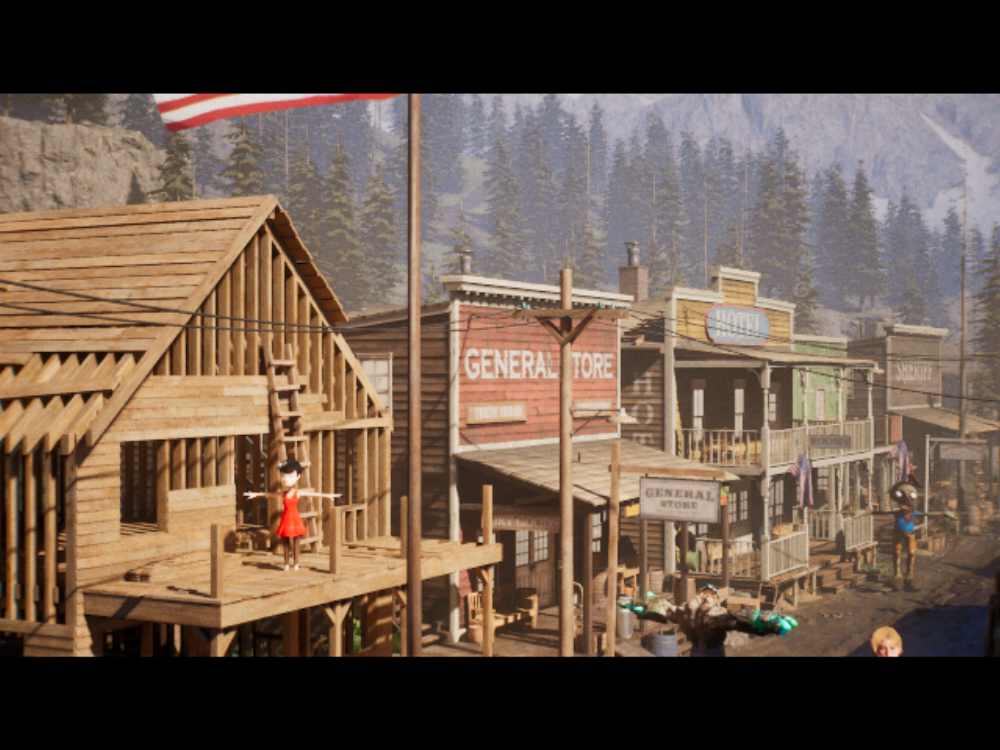}
	  }
	  \subfloat[Redwood Forest]{
		\includegraphics[width=0.245\textwidth]{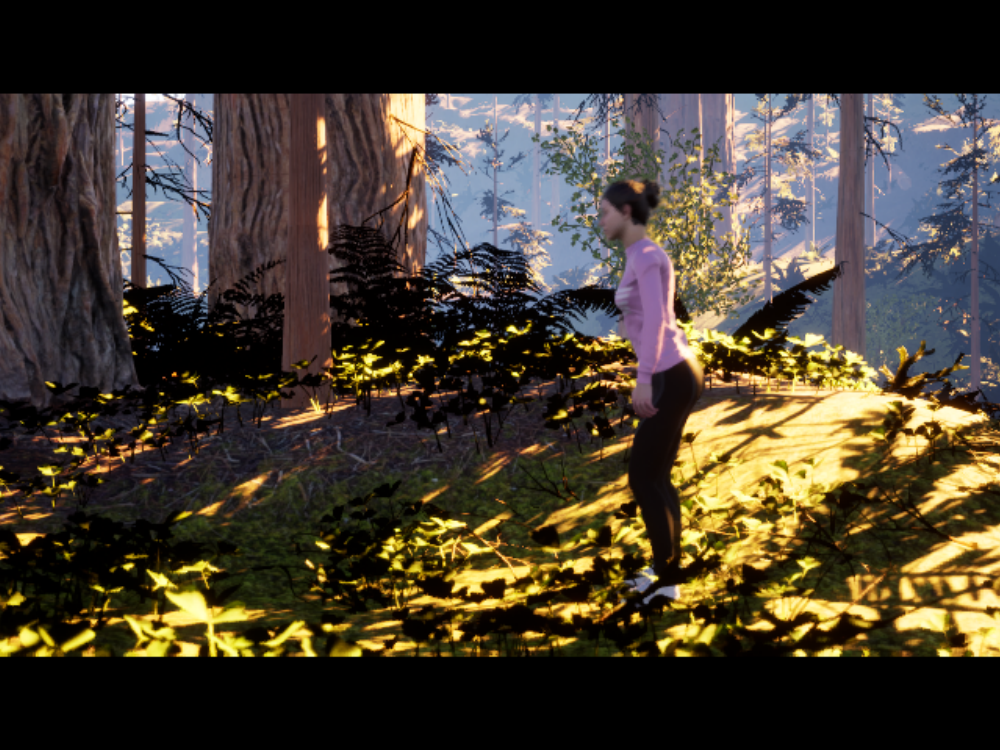}
	  }\\[-2.5ex]
	  \subfloat[Cemetery]{
		\includegraphics[width=0.245\textwidth]{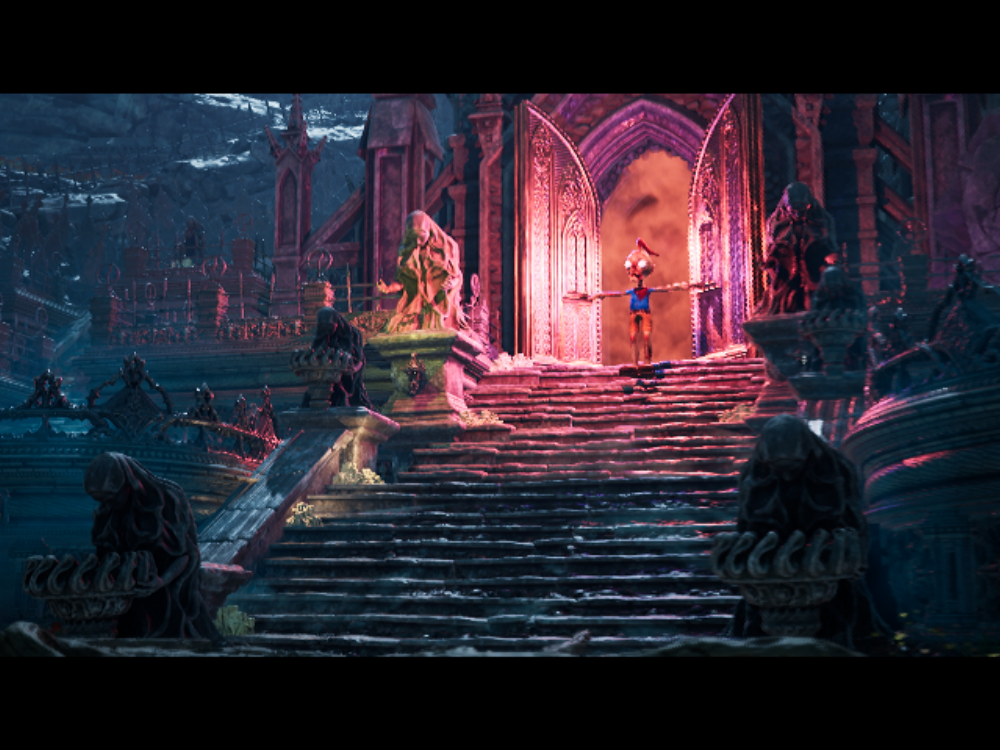}
	  }
	  \subfloat[Bridge]{
		\includegraphics[width=0.245\textwidth]{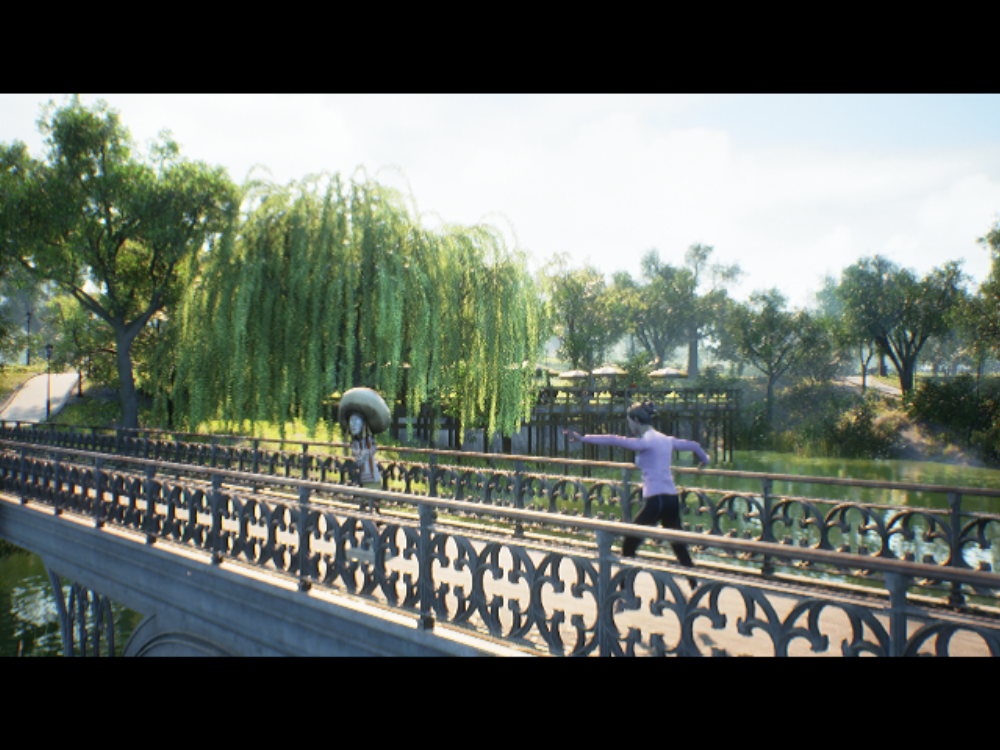}
	  }
	  \subfloat[Downtown West]{
		\includegraphics[width=0.245\textwidth]{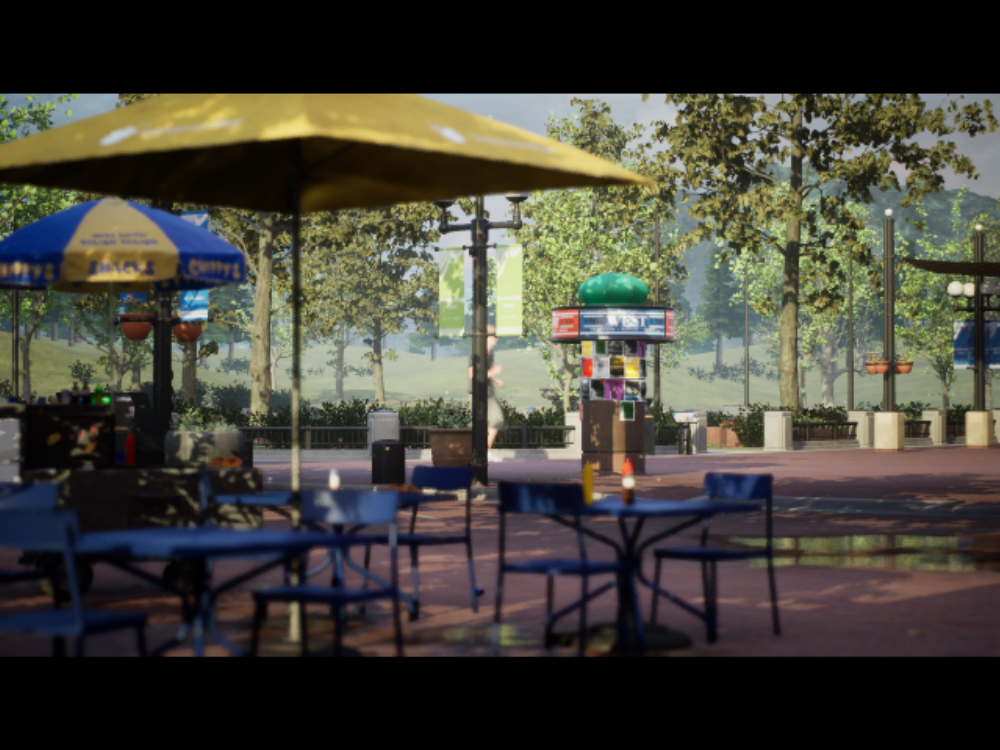}
	  }
	  \subfloat[Tennis Court]{
		\includegraphics[width=0.245\textwidth]{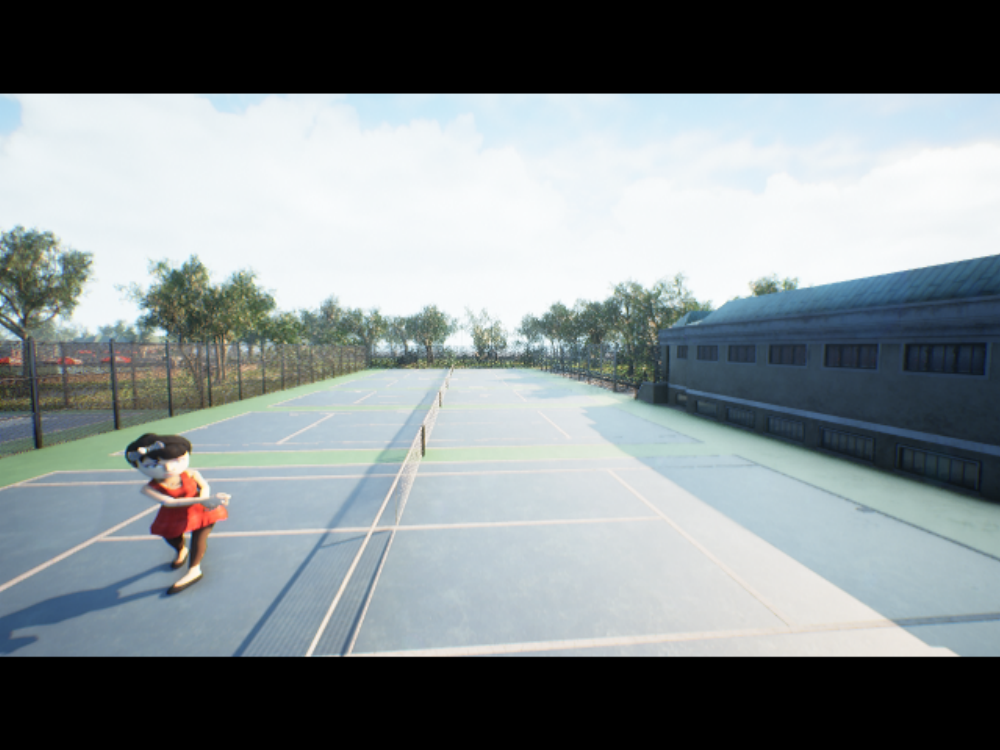}
	  }
	  \caption{Example views from a few sample scenes}
	\label{fig:sample_view}
\end{figure*}

To ensure the generalizability of our approach, we aimed to create a wide-ranging dataset. We gathered 13 background
environments from the UE Marketplace, each with unique artistic styles and complexities. In these environments, we
randomly inserted over 20 different characters along with 30 animation sequences, ranging from simple walks to complex
hip-hop dances. We then selected good viewpoints and created camera paths to follow the main animation character for
each animation sequence to create a variety of sequences. In this regard, we followed a method used to create datasets
in recent works~\cite{mixamo_ref,shugrina2019creative}. Sample scenes of a few applications are shown in
Fig.~\ref{fig:sample_view}.

\begin{table}[]
	\caption{Platform Configuration}
	\footnotesize
	\begin{center}
	  %\resizebox{0.99\columnwidth}{!}{
	  \begin{tabular}{| l l l|}   
		\hline
		%\rowcolor{gray}
		\textbf{Parameter} &
		\multicolumn{2}{l|}{\textbf{Type/Value}} \\ 
		\hline
		%\rowcolor{gray}\multicolumn{3}{c}{Desktop Configuration} \\ \hline
		CPU & \multicolumn{2}{l|}{Intel\textregistered Xeon\textregistered Gold 6226R @ 2.90GHz }\\
		\# CPU cores & \multicolumn{2}{l|}{64}\\
		RAM & \multicolumn{2}{l|}{256 GB} \\
		L1, L2, and L3 cache  & \multicolumn{2}{l|}{ 2 MB, 32 MB, and 44 MB}\\
		GPU &  \multicolumn{2}{l|}{NVIDIA RTX\texttrademark 4080}\\ 
		GPU memory &  \multicolumn{2}{l|}{16 GB}\\
		\#CUDA cores & \multicolumn{2}{l|}{9728}\\ 
		Game engine & \multicolumn{2}{l|}{Unreal Engine v5.1}\\
		\hline
	  \end{tabular}
	  %}
	\end{center}
	\label{tab:config}
\end{table}

\begin{table}[]
	\caption{Graphics benchmarks}
	\footnotesize
	\begin{center}	
	%\resizebox{0.7\textwidth}{!}{
	  \begin{tabular}{|l|l|l|l|l|l|}
		\hline
		%\rowcolor{gray}
	   {\textbf{Abbr.}} &{ \textbf{Name}} &  {\textbf{Res.}} &  \textbf{Graphics} &  \textbf{Engine} & \textbf{Game}    \\
	    & &   &  \textbf{API} & &\textbf{Engine}    \\
		\hline
		 \textit{PR} & City Park & 360p & DX12 & Epic Games & UE   \\
		\hline 
		\textit{WT} & Western Town & 360p & DX12 & Epic Games &UE   \\
		\hline 
		\textit{RF} & Redwood Forest & 360p & DX12 & Epic Games & UE   \\
		\hline
		\textit{CM} & Cemetery & 360p & DX12 & Epic Games &UE   \\
		\hline
		\textit{BR} & Bridge & 360p & DX12 & Epic Games & UE   \\
		\hline
		\textit{DW} & Downtown West & 360p & DX12 & Epic Games & UE   \\
		\hline
		\textit{TC} & Tennis Court & 360p & DX12 & Epic Games & UE   \\
		\hline  
		\textit{LB} & Lab & 360p & DX12 & Epic Games & UE   \\
		\hline 
		\textit{BK} & Bunker & 360p & DX12 & Epic Games & UE   \\
		\hline
		\textit{TR} &  Tropical & 360p & DX12 & Epic Games & UE   \\
		\hline
		\textit{VL} &  Village & 360p & DX12 & Epic Games &UE   \\
		\hline
		\textit{TN} & Town & 360p & DX12 & Epic Games & UE  \\
		\hline
		\textit{SL} &  Slum & 360p & DX12 & Epic Games &UE   \\
		\hline
		\hline
		\multicolumn{6}{|l|}{UE: Unreal Engine, DX: DirectX} \\
		\hline
	  \end{tabular}
	  % }
	 \end{center} 
	\label{tab:bench}
\end{table} 
	
\subsection{Variation in the Frame Rate}

\label{subsec:rendering}
\begin{figure}[!h]
	\centering
	\includegraphics[width=0.99\columnwidth]{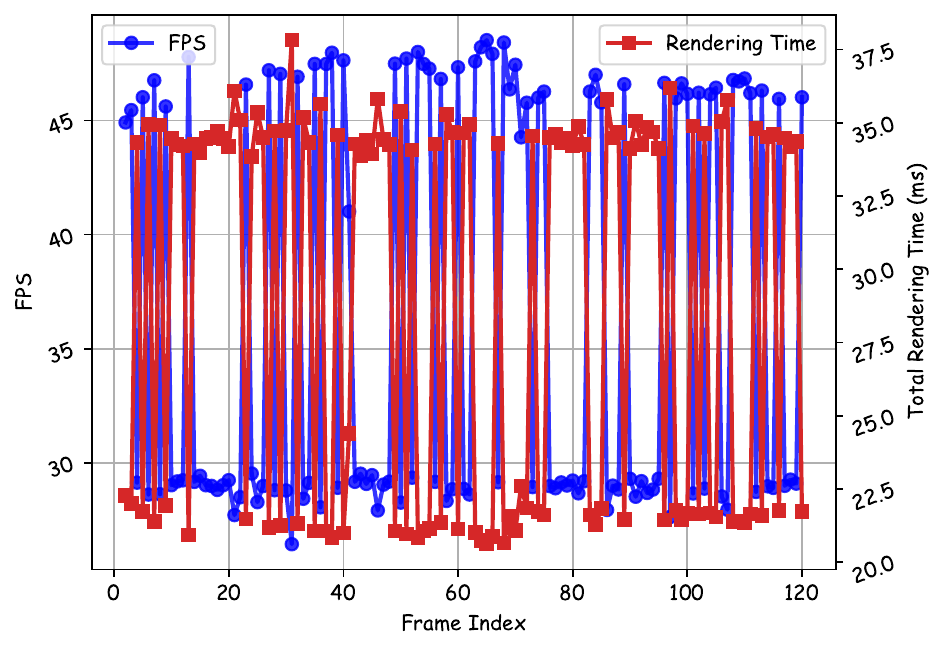}
	\caption{Variation in the total rendering time}
	\label{fig:total}
\end{figure}

As highlighted in Section~\ref{sec:Introduction}, visual artifacts such as screen tearing and judder can occur even with
a high refresh rate display. This is often due to the irregular delivery of frames from the GPU. This section delves
deeper into the variability of frame rates in real-world applications. We conducted extensive experiments, measuring the
total rendering time for each frame in a scene with a single dynamic object. By plotting these values over time (see
Fig.~\ref{fig:total}), we observe significant fluctuations in the FPS (frames per second) with some frames taking
considerably longer to render. The average FPS is almost 29, while the standard deviation is 6.6, indicating that the
FPS values show considerable variability or spread around the mean. This inconsistency leads to noticeable flickering,
distracting viewers and degrading image quality. Our findings underscore the necessity of temporal supersampling to
stabilize the frame rate, aligning it more closely with the display's refresh rate, and thus ensuring high-quality,
flicker-free images. 

\begin{figure}[!h]
	\centering
	\includegraphics[width=0.99\columnwidth]{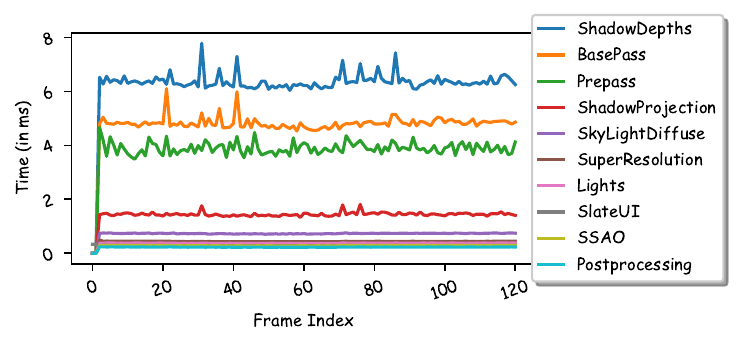}
	\caption{Top 10 high-latency steps in the rendering process}
	\label{fig:topsteps}
\end{figure}

When a frame is being rendered, it passes through a series of steps that form the rendering pipeline. To better understand
the reasons behind the high rendering time and variability, we conduct a detailed analysis of the pipeline. We
identify the top ten high-latency steps that cause delays and plot them in Fig.~\ref{fig:topsteps}. The plot shows that
the ShadowDepths, BasePass, PrePass and ShadowProjections steps are the most time-consuming ones. These steps involve
complex calculations and require significant computing resources, which can result in high rendering times.

% \noindent
% \resizebox{0.99\columnwidth}{!} {
% \begin{tcolorbox}[colback=gray!10]
% \parbox{\columnwidth}{
% \textbf{For a real-world application, we achieve an average FPS of 29 with a standard deviation of 6.6. This highlights the necessity of using temporal supersampling to ensure a stable frame rate.}}
% \end{tcolorbox}
% }

\begin{mybox}[label=box:frame_var]{Insight}
	For a real-world application, we achieve an average FPS of 29 with a standard deviation of 6.6. This highlights the necessity of using temporal supersampling to ensure a stable frame rate.
\end{mybox}
	
\subsection{Foveated Segmentation}
\label{subsec:fov_seg}

\begin{figure}[!h]
	\centering
	\includegraphics[width=0.99\columnwidth]{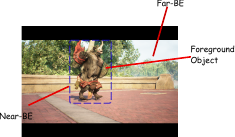}
	\caption{Types of objects in a frame}
	\label{fig:be}
\end{figure}

As mentioned in Section~\ref{subsec:foreground_bias}, the foreground bias effect can be used to our advantage by
extrapolating foreground interactions more efficiently since artifacts are more noticeable in the foreground as compared to
the background. To achieve this, we segment a frame into two parts: foreground and background. This is known as foveated
segmentation~\cite{bjorkman2005foveated}. We can then use different extrapolation algorithms for each part and blend the
outputs. However, there is another effect that complicates this simple division of the frame. This effect is known
as the near-object effect~\cite{meng2020coterie}, which means that changes in the pixels in the background around the
foreground objects are more noticeable than those further away from the foreground objects. Hence, similar to
Coterie~\cite{meng2020coterie}, we, too, divide the entire frame into three regions: FI (foreground interactions),
Near-BE (near background environment) and Far-BE (refer Fig.~\ref{fig:be}). We propose a {\em novel algorithm} in this
space. The near and far background environments
are separated by a rectangular boundary of width and height, $w$ and $h$, respectively. In
Section~\ref{subsec:segmentation}, we will delve into the selection process for determining the width ($w$) and height
($h$).

\begin{mybox}[label=box:fov_seg]{Insight}
	The extrapolation can be more efficiently performed by exploiting the fact that humans perceive different parts of an image with varying levels of sensitivity.
\end{mybox}
\begin{table}[!htbp]
    \centering
	\footnotesize
	\resizebox{0.99\columnwidth}{!}{
    \begin{tabular}{|l|l|l|l|}
        \hline
        \textbf{Aspect Ratio} & \textbf{Resolution} & \textbf{Name} & \textbf{\# Pixels} \\
        \hline
        \multirow{8}{*}{4:3} & 320 x 240   & QVGA & 0.0768 \\
                             & 640 x 480   & VGA & 0.3072 \\
                             & 800 x 600   & SVGA & 0.4800 \\
                             & 1024 x 768  & XGA & 0.7864 \\
                             & 1280 x 960  & SXGA- & 1.2288 \\
                             & 1400 x 1050 & SXGA+ & 1.4700 \\
                             & 1600 x 1200 & UXGA & 1.9200 \\
                             & 2048 x 1536 & QXGA & 3.1457 \\
        \hline
        \multirow{9}{*}{16:9} & 640 x 360   & nHD & 0.2304 \\
                              & 854 x 480   & FWVGA & 0.4099 \\
                              & 1280 x 720  & HD or 720p & 0.9216 \\
                              & 1600 x 900  & HD+ & 1.4400 \\
                              & 1920 x 1080 & Full HD or 1080p & 2.0736 \\
                              & 2560 x 1440 & Quad HD or 1440p & 3.6864 \\
                              & 3840 x 2160 & 4K or UHD & 8.2944 \\
        \hline
    \end{tabular}
	}
    \caption{Common resolutions for 4:3 and 16:9 aspect ratios with pixel counts (in millions)}
    \label{tab:resolutions}
\end{table}

\begin{figure}[]
	\centering
	\includegraphics[width=0.99\columnwidth]{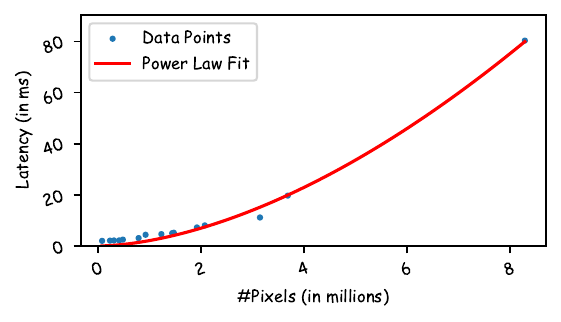}
	\caption{Relation between the frame size and the inpainting latency}
	\label{fig:rel}
\end{figure}

\subsection{Latency of the Inpainting Task}
As mentioned in Section~\ref{sec:Introduction}, we propose to segment the warped frame into three distinct patches. Each
patch is subsequently inpainted in parallel by separate neural networks. This approach leverages the benefits of
parallel processing, which is advantageous from the point of view of performance, especially 
when the task execution time is a
superlinear function of the task size. In other words, if the inpainting latency increases super-linearly with
the input frame size, we can naturally justify frame splitting and parallelization. 

To illustrate this, we inpaint frames rendered at various resolutions (refer to Table~\ref{tab:resolutions}) and
compute their latencies. The results, shown in Figure~\ref{fig:rel}, indicate that the relationship between the frame
size (total number of pixels) and the latency is not linear. Instead, the data indicates a power-law relationship ($a
\cdot x^b$), confirming that the inpainting latency is indeed a superlinear function of the input frame size.

\begin{mybox}[label=box:lat]{Insight}
	The inpainting task exhibits a superlinear increase in latency relative to the input frame size, making it suitable for partitioning and parallel execution.
\end{mybox}

\subsection{Frame Segmentation}
\label{subsec:math_prop}
In Section~\ref{subsec:fov_seg}, we discussed the partitioning of a frame based on how the human eye perceives distinct
regions and its sensitivity to subtle variations within these regions. Building upon this foundation, in this section,
we delve into the mathematical properties of these regions. We shall perform a comprehensive characterization of the rendered
frames, analyzing their properties in the temporal domain~\cite{bouwmans2018role, li2004statistical}. This is a highly novel
characterization approach for real-time graphics systems, such as Virtual Reality (VR). This novel inter-frame
analysis is a basis for proposing a heuristic-based approach to distinguish and segment frames into the
foreground, near-background, and far-background regions. 

In the temporal domain, we analyze dynamic changes over time by capturing inter-frame variations in the pixel intensity.
Mathematically, for a pixel at position $(x, y)$ in frame $F$, the temporal change $T(x,y)$ can be calculated using
Equation~\ref{eq:temp_var}. After computing the variation values, we apply Otsu's thresholding
method~\cite{liu2009otsu} to divide the frame into two primary regions: near and far. Pixels exhibiting high temporal
variation are classified as part of the near region. This high variation indicates significant movement or changes in
the scene. Conversely, pixels with low variations are classified as part of the far region. These areas tend to have
more stable intensity values over time. Note that these regions can be spatially disconnected.

\begin{subequations}
		\renewcommand{\theequation}{\theparentequation.\arabic{equation}}
		\captionsetup{labelformat=parens,labelsep=space}
		\begin{equation}
			T(x, y) = \frac{1}{T-1} \sum_{t=1}^{T-1} |I(x, y, t+1) - I(x, y, t)|
			\label{eq:temp_var}
		\end{equation}
		\noindent 	
		where, $I(x,y,t)$ represents the intensity of the pixel in the frame rendered at time $t$ and $N$ is the total number of frames. Higher values of $T(x,y)$ indicate greater temporal variability.
\end{subequations}

\begin{figure*}[!h]
    \subfloat[]{
	    \includegraphics[width=0.249\textwidth]{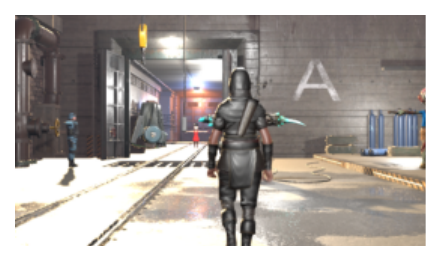}
	    \label{fig:bk3_orig}
    }
    \subfloat[]{
	    \includegraphics[width=0.249\textwidth]{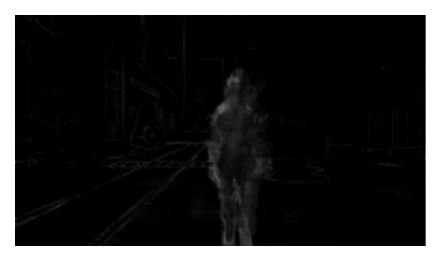}
	    \label{fig:bk3_temp_var}
    }
	\subfloat[]{
	    \includegraphics[width=0.249\textwidth]{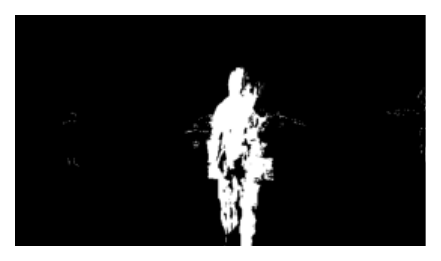}
	    \label{fig:bk3_high_mask}
    }
	\subfloat[]{
	    \includegraphics[width=0.249\textwidth]{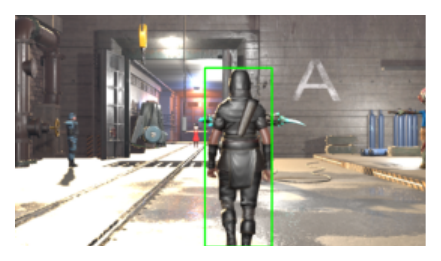}
	    \label{fig:bk3_temp_orig}
    }\\[-2.5ex]
	\subfloat[]{
	    \includegraphics[width=0.249\textwidth]{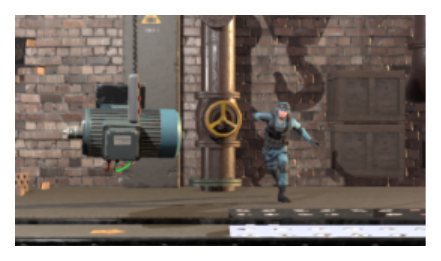}
	    \label{fig:bk7_orig}
    }
    \subfloat[]{
	    \includegraphics[width=0.249\textwidth]{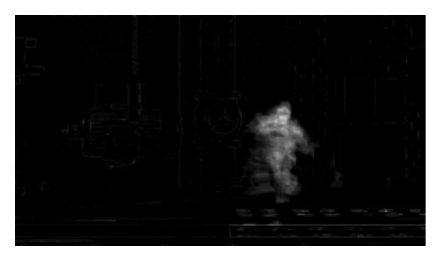}
	    \label{fig:bk7_temp_var}
    }
	\subfloat[]{
	    \includegraphics[width=0.249\textwidth]{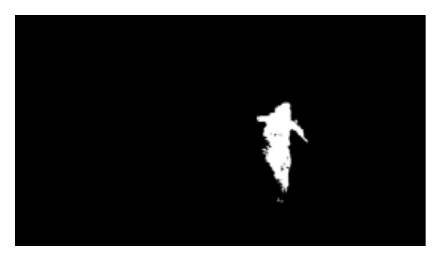}
	    \label{fig:bk7_high_mask}
    }
	\subfloat[]{
	    \includegraphics[width=0.249\textwidth]{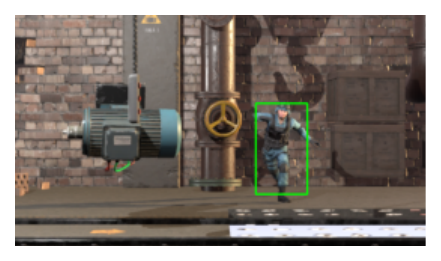}
	    \label{fig:bk7_temp_orig}
    }
	\\[-2.5ex]
	\subfloat[]{
	    \includegraphics[width=0.249\textwidth]{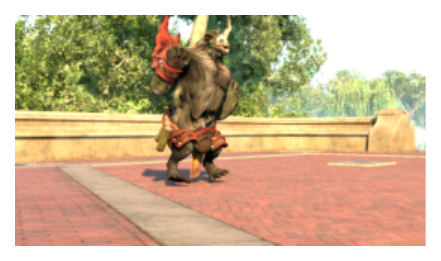}
	    \label{fig:pr1_orig}
    }
    \subfloat[]{
	    \includegraphics[width=0.249\textwidth]{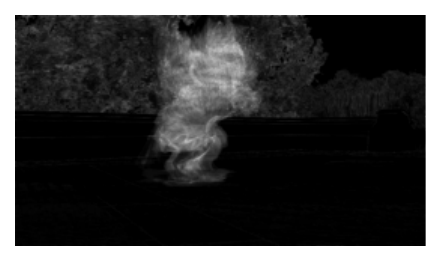}
	    \label{fig:pr1_temp_var}
    }
	\subfloat[]{
	    \includegraphics[width=0.249\textwidth]{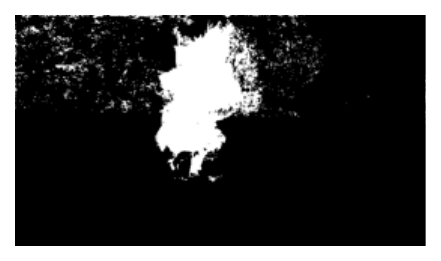}
	    \label{fig:pr1_high_mask}
    }
	\subfloat[]{
	    \includegraphics[width=0.249\textwidth]{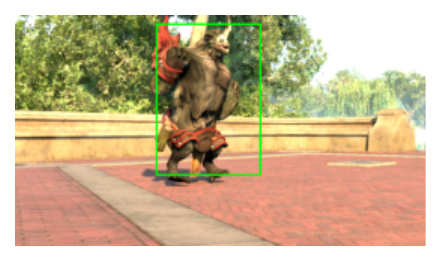}
	    \label{fig:pr1_temp_orig}
    }
    \caption{(BK: \protect\subref{fig:bk3_orig} Original frame; \protect\subref{fig:bk3_temp_var} Temporal variations; \protect\subref{fig:bk3_high_mask} High variations mask; \protect\subref{fig:bk3_temp_orig} Rectangular boundary for the near region); (BK: \protect\subref{fig:bk7_orig} Original frame; \protect\subref{fig:bk7_temp_var} Temporal variations; \protect\subref{fig:bk7_high_mask} High variations mask; \protect\subref{fig:bk7_temp_orig} Rectangular boundary for the near region); (PR: \protect\subref{fig:pr1_orig} Original frame; \protect\subref{fig:pr1_temp_var} Temporal variations; \protect\subref{fig:pr1_high_mask} High variations mask; \protect\subref{fig:pr1_temp_orig} Rectangular boundary for the near region).}
    \label{fig:inter_frame}  
\end{figure*}

We present the segmentation results for three benchmark scenes in Figure~\ref{fig:inter_frame}. For each scene, we
include the following visualizations: the original frame, computed temporal variations, a binary mask highlighting
regions with high variations and a rectangle fitted around the high temporal variation regions, superimposed on the
original frame. We make the following observations from the figures:

\circled{1} Dynamic objects consistently appear in the regions of high temporal variations. This includes not only the
moving object itself but also its immediate surroundings, which are affected by the object's movement.

\begin{figure}[!h]
	\centering
	\includegraphics[width=0.99\columnwidth]{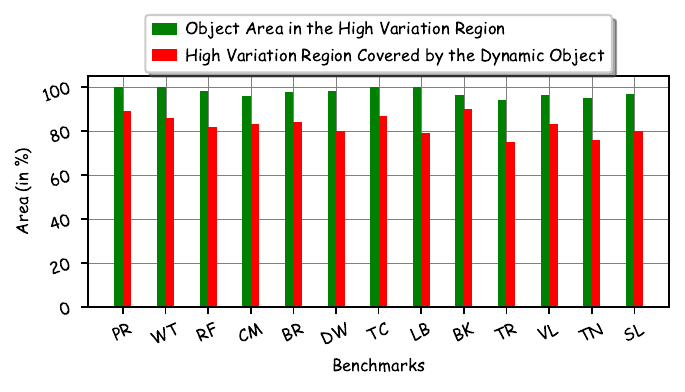}
	\caption{Analysis of the high variation regions and coverage of dynamic objects}
	\label{fig:area}
\end{figure}

To substantiate our findings across benchmarks, we further analyze the relationship between the dynamic objects and the
high temporal variation regions. We plot the percentage of the dynamic objects' area that lies within the high variation
regions along with the area of the high variation regions that is occupied by the dynamic objects (refer to
Figure~\ref{fig:area}). The average coverage of the high variation regions by the dynamic objects across all benchmark
scenes is approximately 96.5\%. This indicates that the majority of the high variation regions are occupied by the dynamic
objects. 82.6\% of the dynamic objects' area falls within the high variation regions. This high percentage
demonstrates that the maximum temporal variation is in the regions where dynamic objects are present, with a small 
amount of exclusion.

%FIXME: plural in area.pdf

\begin{figure}[!h]
	\centering
	\includegraphics[width=0.99\columnwidth]{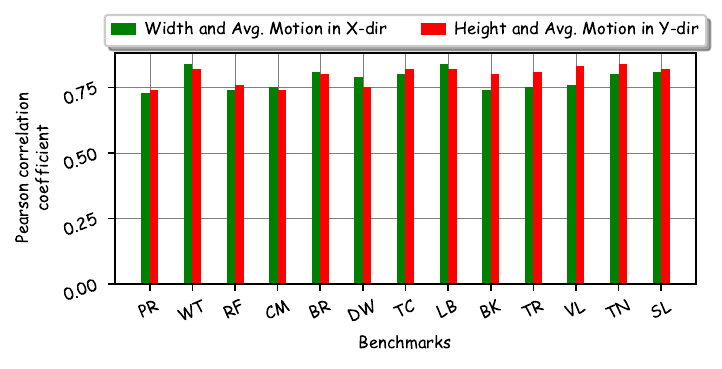}
	\caption{Relationship between boundary dimensions and average motion}
	\label{fig:corr}
\end{figure}

In order to find the rectangular borders of the foreground, near-BE and far-BE regions
in real time, we have developed a heuristic-based method. This approach is necessary
because it is not feasible to use future frames for analyzing variations between frames in real time. To facilitate this
heuristic approach, we analyze the data to detect patterns and correlations between the dimensions of the rectangular
boundaries and the average motion. For each frame, we measure the average motion in the x-direction ($\mu_x$) and the
y-direction ($\mu_y$) and the corresponding width ($w$) and height ($h$) of the rectangular boundaries for the regions with
high temporal variation. Then, we calculate the Pearson correlation coefficient for $w$ and $\mu_x$, and for $h$ and
$\mu_y$ (see Figure~\ref{fig:corr}). The correlation coefficient helps us quantify the strength and direction of the
linear relationship between these variables. We make the following observations from the results:

\circled{1} For a given dynamic object, the width of the rectangular boundary 
has a strong positive correlation with the average motion in the
x-direction ($\mu_x$). The average Pearson's correlation coefficient is 0.78. Additionally, the height of the
rectangular boundary has a strong positive correlation with the average motion in the y-direction ($\mu_y$), with an
average correlation coefficient of 0.79. 

\circled{2} These strong correlations provide a solid foundation for our heuristic approach for real-time frame
segmentation. By leveraging the linear relationships between boundary dimensions and motion, we can estimate the
rectangular boundary in real-time without needing future frame data.

\begin{mybox}[label=box:inter_frame]{Insight}
	By analyzing the motion patterns of dynamic objects, we can effectively segment a frame into foreground, near-BE and far-BE regions. 
\end{mybox}

\subsection{Performance of Various Warping Methods}
\label{subsec:perf_warp}

\begin{figure*}[!h]
	\centering
	\includegraphics[width=0.99\textwidth]{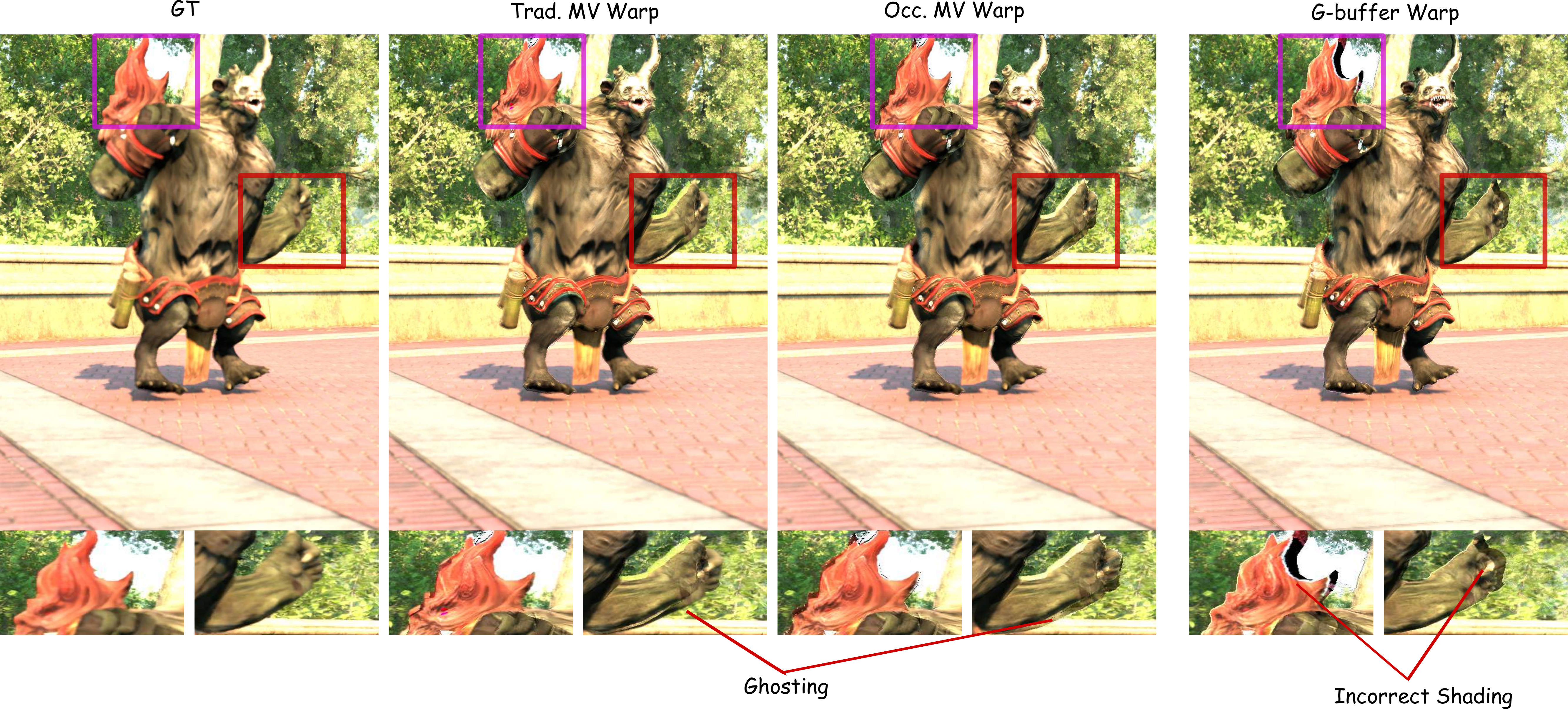}
	\caption{Comparison of various warping techniques}
	\label{fig:warp}
\end{figure*}

In Fig.~\ref{fig:warp}, we present the results of various warping techniques including traditional motion vector-based
warping, occlusion motion vector-based warping and G-buffer-based warping, in comparison to the ground truth. The
results demonstrate that both motion vector-based warping methods exhibit some degree of ghosting, which affects the
overall quality of the output. However, in the case of G-buffer-based warping, ghosting is absent but it leads to
incorrect shading in the resulting frame. Ghosting is absent here because the knowledge of disoccluded objects
is present in the G-Buffers and this information can be used to fill in the gaps.
Therefore, our approach utilizes G-buffer-based warping as an initial
prediction. This addresses the first challenge to some extent (much more needs to be done).

\subsection{Challenges in Frame Extrapolation}
\label{subsec:challenges}

Our primary objective is to predict an intermediate frame called $F_{t+0.5}$ by utilizing the previously rendered frame,
$F_t$. We can achieve this goal due to two primary reasons. First, there are similarities between frames in any
graphics application. Second, we have some internal information from the rendering pipeline, such as motion vectors
that indicate the position of the pixels in the next frame. 

\begin{figure}[!h]
	\centering
	\includegraphics[width=0.99\columnwidth]{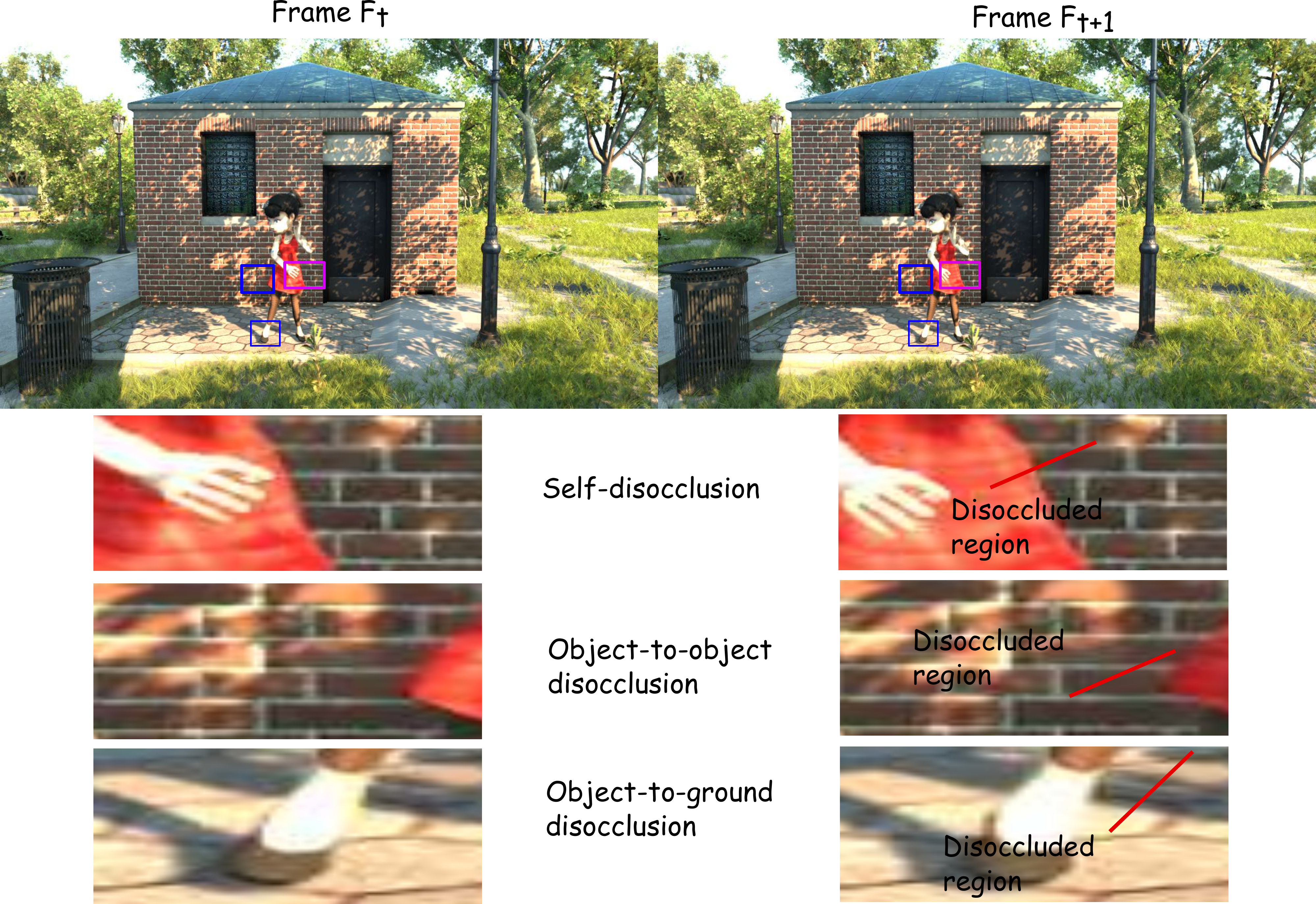}
	\caption{Possible disoccluded regions in a frame}
	\label{fig:disocclusion}
\end{figure}

However, this task is not as simple as it seems, and there are two significant challenges with motion-vector-based
approaches. Since we are extrapolating, we only have motion vectors in one direction, which implies that we can
accurately predict regions that were present in both the frames $F_t$ and $F_{t+1}$. However, for the disoccluded
regions in frame $F_{t+1}$, the motion vectors do not provide any information, and we don't know how to fill those
pixels correctly -- this makes the extrapolation task extremely challenging. Furthermore, there are three categories of
occlusions: self, object-to-object and object-to-background, which also means that three types of disoccluded regions
exist (refer to Fig.~\ref{fig:disocclusion}). Self-occlusion occurs when an object obstructs itself in the image.
Object-to-object occlusion happens when two or more objects overlap, and object-to-background occlusion occurs when an
object is partially or wholly occluded by the background. These disoccluded regions remain a challenge to fill in,
making the extrapolation task even more complex. 

\begin{figure}[!h]
	\centering
	\subfloat[Frame $F_t$]{
	  \includegraphics[width=0.5\columnwidth]{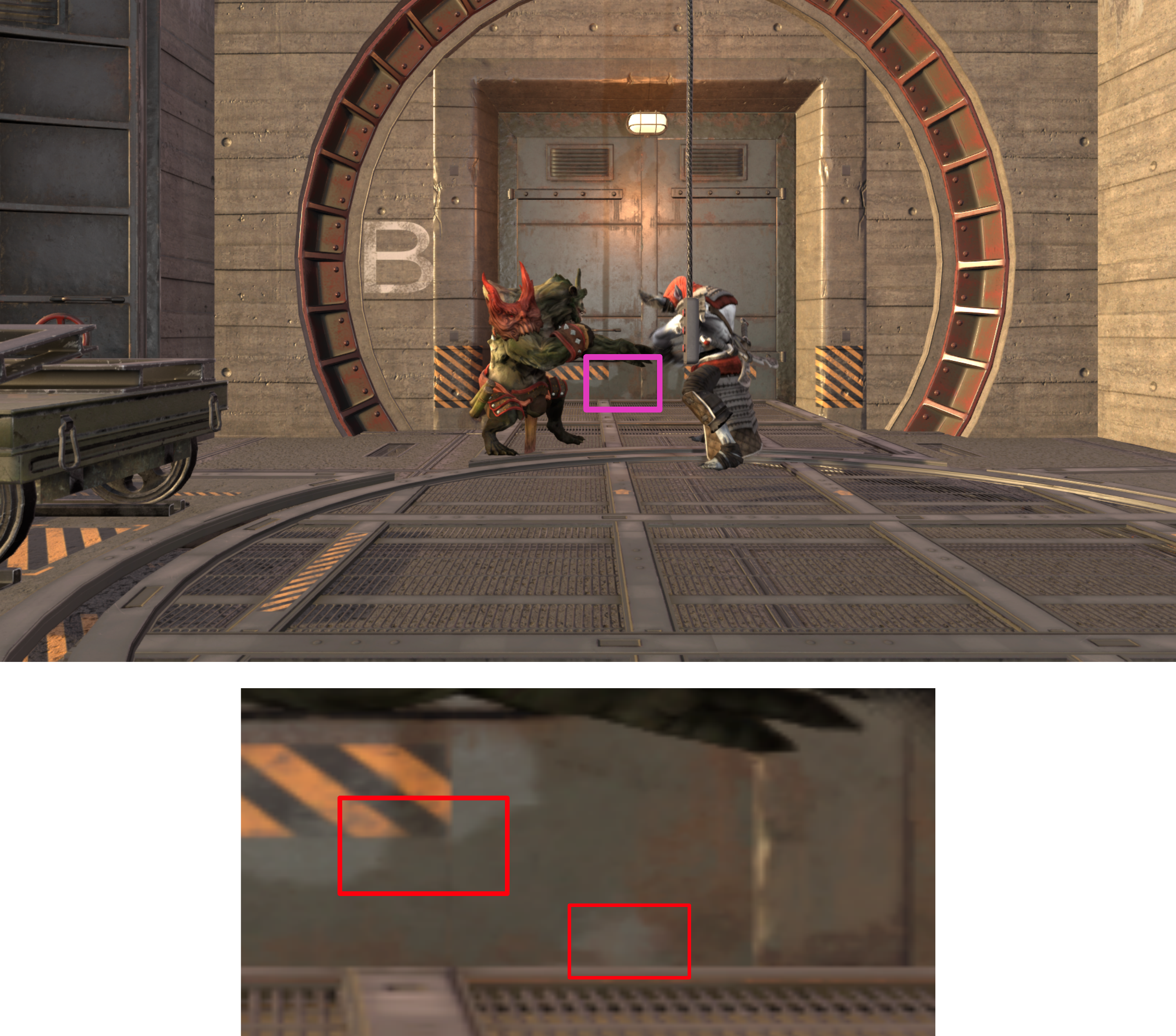}
	}
	\subfloat[Frame $F_{t+1}$]{
	  \includegraphics[width=0.5\columnwidth]{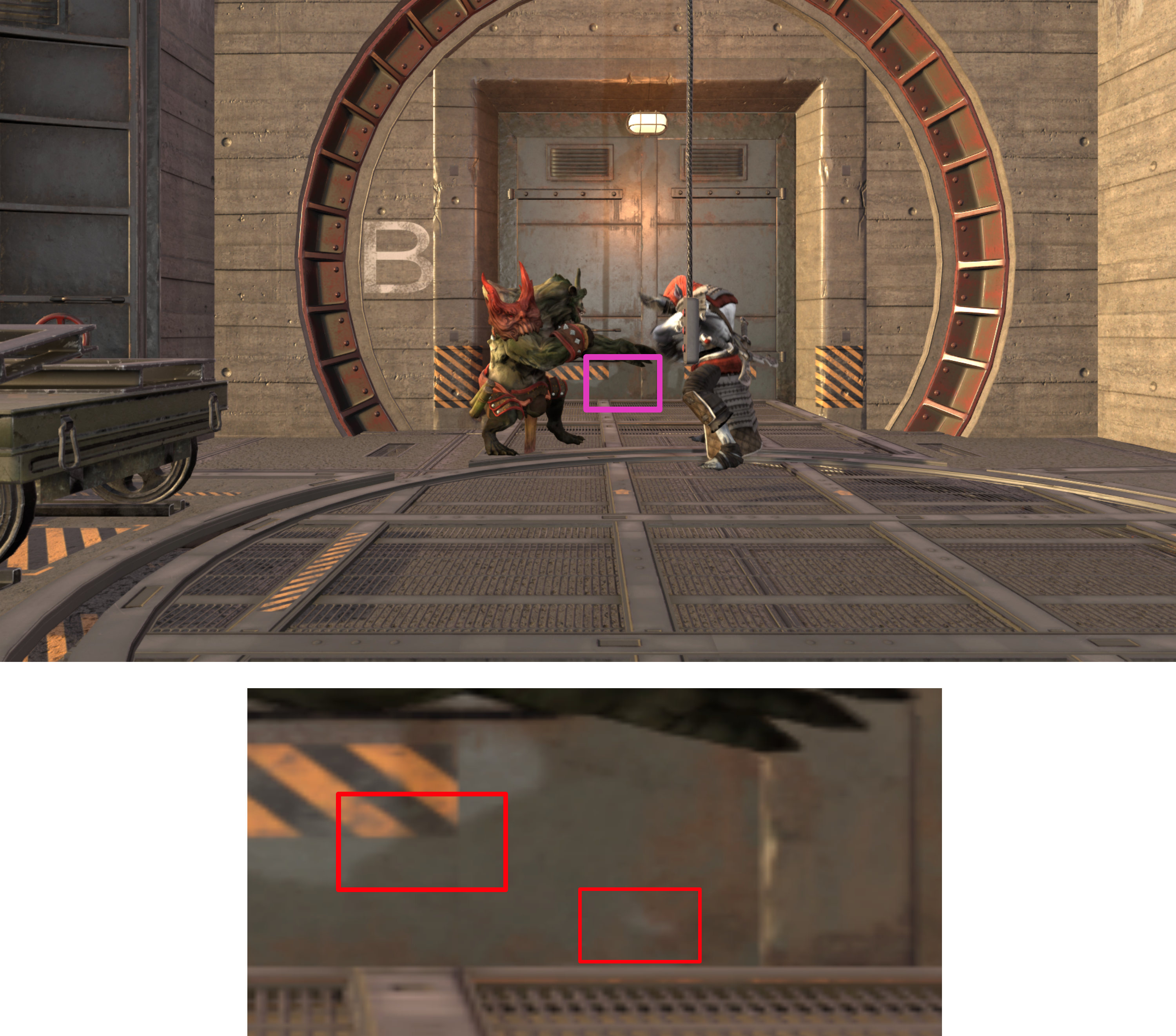}
	}
	\caption{Dynamic changes in the appearance of shadows between two successive frames, $F_t$ and $F_{t+1}$.}\label{fig:shadow}
\end{figure}

The other challenge in the extrapolation task is accurately predicting changes in {\em shadows}. Even minor changes in the
movement of a dynamic object can result in significant changes in the shadow it casts, as illustrated in
Fig.~\ref{fig:shadow}. This can have a significant impact on the overall realism of graphics applications, as shadows
play a crucial role in conveying depth and dimensionality.
%FIXME: Highlight these two challenges with graphical elements better

To tackle these two challenges, we divide the extrapolation task into two parts. The first part focuses on correctly
extrapolating the frame except for the shadow. This involves predicting the movement and transformations of objects in
the scene while ignoring changes in the shadow. The second part deals exclusively with predicting the shadow. This
involves accurately predicting how the shape, size, and intensity of the shadow will change over time as the object
moves and the lighting conditions in the scene change. 

\begin{mybox}[label=box:challenges]{Insights}
	\circled{1} The presence of disoccluded regions (\ding{112}) and sudden changes in the shadows (\tikz\draw[fill=black!20!gray] (0,0) ellipse (0.75ex and 0.5ex);) pose a significant challenge for
	frame extrapolation.
	
	\circled{2} We propose two segmentation approaches: first, separating shadows from the rendered frame, and second,
	divide the rendered frame (without shadows) into three patches (foreground, near-BE and far-BE) by leveraging foreground bias and near-object
	effects. These segmentation techniques aim to address challenges associated with frame extrapolation.
\end{mybox}

% \subsection{Impact of Latency on User Experience}
% As mentioned in Section~\ref{subsubsec:inter_vs_extra}, interpolation introduces additional latency, significantly
% affecting the user experience due to the human visual system's acute sensitivity to delays. Even minor delays can
% significantly disrupt the viewing experience. The concept of the just noticeable delay (JND) underscores this
% sensitivity, indicating that humans can normally detect delays as low as 3-5 ms with the threshold for gamers and active
% young people being even lower. It needs to be less than 1 ms in the case of Head-Mounted Displays
% (HMDs)~\cite{jerald-thesis, 3ms,haptics}. Given these factors, the latency introduced by interpolation can easily exceed
% the JND thresholds, leading to perceptible delays and compromising the quality of the viewing experience.
%FIXME: This should go where the interpolation section is. Highlight it to look like a motivation section as well. 

%% file: overview.tex
\section{Methodology}
\label{sec:methodology}

\begin{figure*}[!h]
	\centering
	\includegraphics[width=0.99\textwidth]{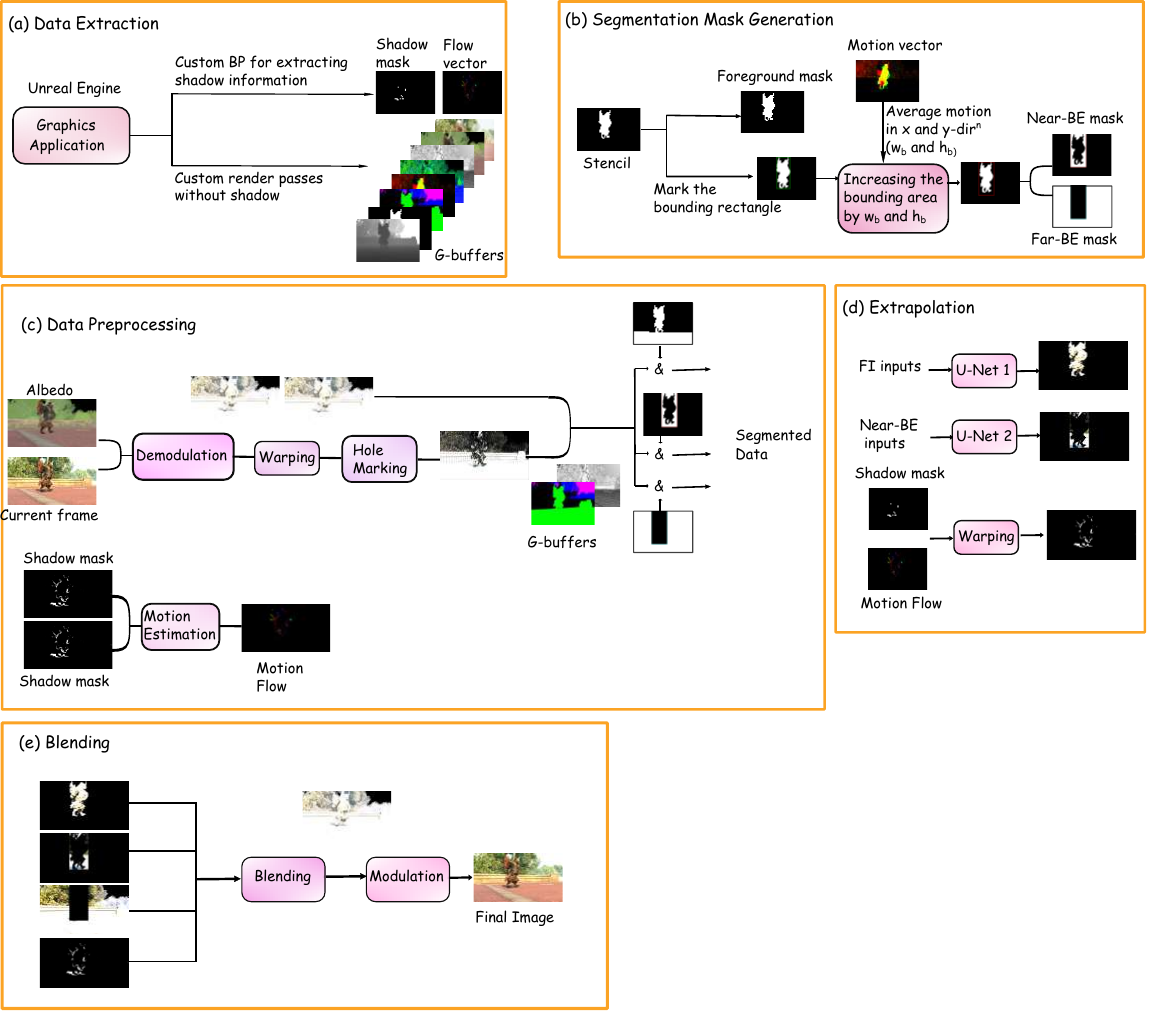}
	% \captionsetup{width=\columnwidth}
	\caption{ Overview of our proposed approach.(a) \textit{Data Extraction}: We extract a few G-buffers using custom render passes and obtain the shadow information using a custom blueprint (BP) class from the Unreal Engine (UE). (b) \textit{Segmentation}: We create three binary masks to segment the frames into the foreground, near-BE, and far-BE regions. (c) \textit{Data Preprocessing}: Using the extracted G-buffers, we warp $F_t$, mark invalid pixels or holes in the warped frame, and segment the warped frame into three distinct regions. For shadows, we compute the motion flow. (d) \textit{Extrapolation}: We use two different neural networks to extrapolate the foreground and near-BE regions, while the warped frame is used for the far-BE region. For shadow extrapolation, we employ occlusion motion-vector-based warping. (e) \textit{Blending}: Finally, we merge the three extrapolated regions and the shadows to produce the final image.}
		\label{fig:overview}
\end{figure*}

In this section, we establish a formal definition of the frame extrapolation problem before detailing our methodology. Our primary objective is to predict the intermediate frame $F_{t+0.5}$ based on the preceding frame $F_t$ and a few G-buffers. In simple terms, we aim to generate a frame that visually sits halfway between two consecutive frames while ensuring coherence with the overall sequence of frames. To achieve this, our method is organized into five separate stages as shown in Fig.~\ref{fig:overview}. These five stages are as follows:

\begin{enumerate}
\item \textbf{Data Extraction:} The first stage is data extraction, where we run the applications on the Unreal Engine. In this phase, we
extract various G-buffers utilizing custom render passes. Subsequently, we preprocess the extracted data to ensure its
compatibility with the subsequent stages. In addition to the G-buffers, we extract the shadow information from the
rendering engine.

\item \textbf{Segmentation mask generation:} In this stage, we create three binary masks that partition the
rendered frame into three distinct patches: foreground, near-background (near-BE), and far-background (far-BE). As
discussed in Section~\ref{subsec:fov_seg}, only dynamic objects in the frame are identified as the foreground, while the
rest of the screen area is considered as the background. Hence, the stencil G-buffer, which serves as a mask for dynamic
objects is assigned as the foreground mask (refer Fig.\ref{fig:buffers}). Next, to find the near-BE region, we initially
define a bounding rectangle around dynamic objects and subsequently expand its dimensions by small increments determined
by the motion vector. This expanded bounding rectangle is then assigned as the mask for the near-BE region, while the
residual area constitutes the far-BE mask.

%FIXME: Instead of first, second, third have numbers. Makes it look more formal.
\item \textbf{Data preprocessing:} Here, we perform two critical tasks. First, we execute G-buffer-guided
warping and then identify {\em invalid pixels} in the warped frame. Subsequently, we perform a bitwise \textit{AND} operation
between the three segmentation masks and the warped frame to create three distinct input sets for extrapolation. For
shadows, we utilize the shadow mask extracted from Frame $F_t$ to compute the motion vectors. 

\item \textbf{Extrapolation:} As outlined in Section~\ref{sec:Introduction}, we employ two different bespoke neural networks to
extrapolate the foreground and near-BE regions. For the far-BE region, we retain the warped frame as it is.
Concurrently, we extrapolate shadows using the motion vectors computed in the previous stage. 

\item \textbf{Blending:} This stage involves merging the outputs from the three extrapolation methods along with the predicted shadow.
This integrated approach ensures an accurate extrapolated frame.
\end{enumerate}

In the following section, we will provide a detailed discussion of our proposed method.

%% file: model.tex
\section{Implementation}
\label{sec:Implementation}

\subsection{Data Extraction}
\label{subsec:data}

\begin{figure*}[!h]

    \subfloat[]{
	    \includegraphics[width=0.25\textwidth]{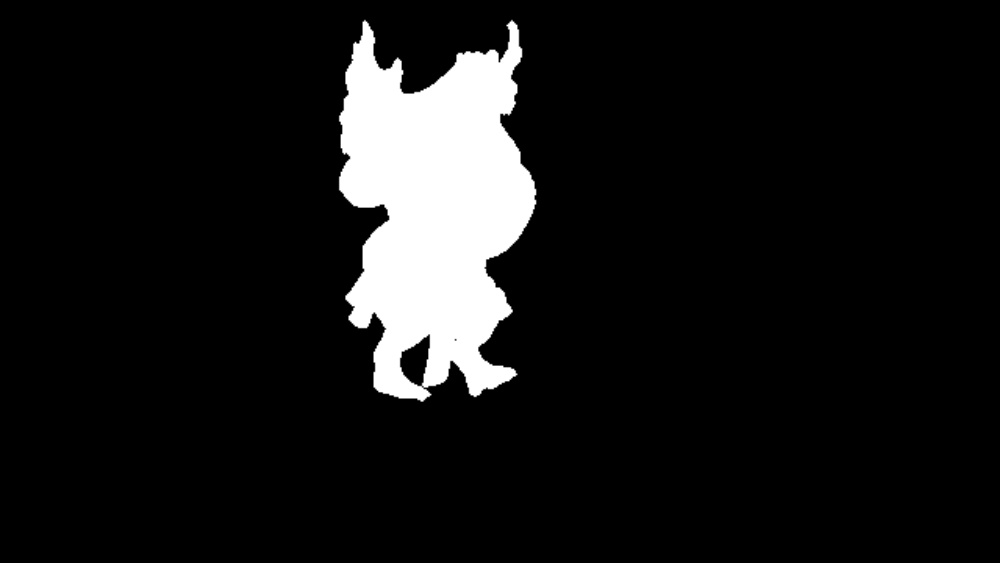}
	    \label{fig:stencil}
    }
    \subfloat[]{
	    \includegraphics[width=0.25\textwidth]{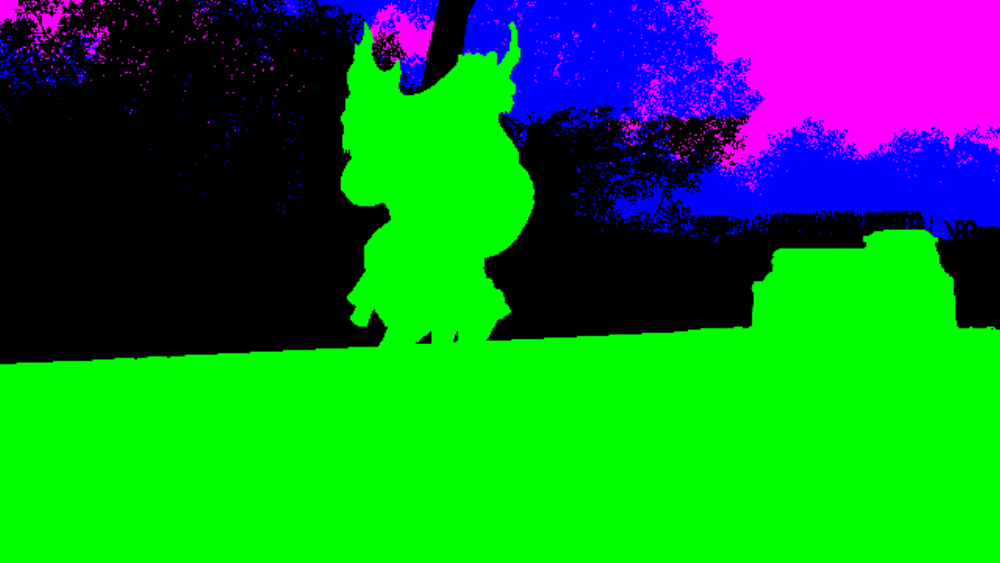}
	    \label{fig:worldnormal}
    }
    \subfloat[]{
	    \includegraphics[width=0.25\textwidth]{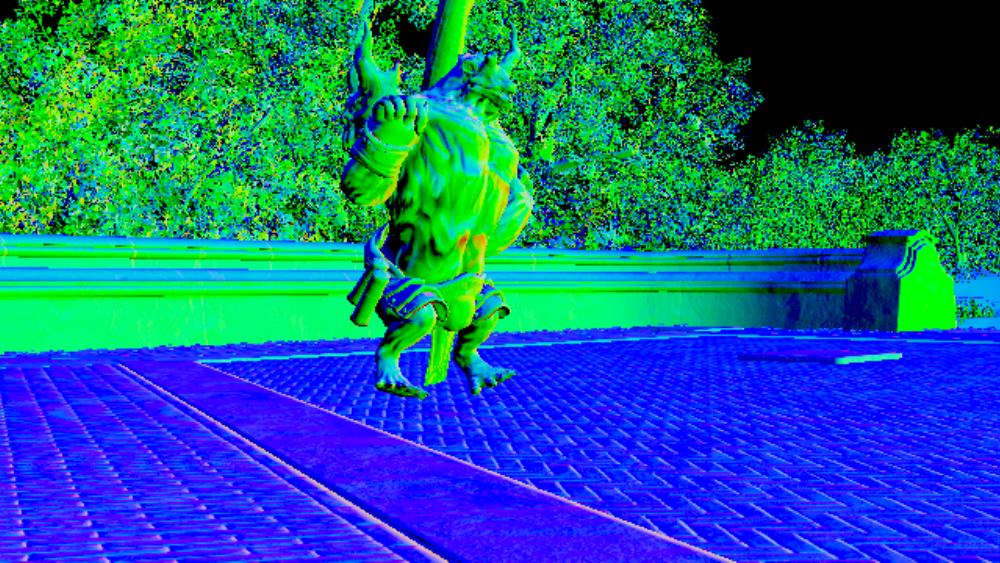}
	    \label{fig:worldposition}
    }
    \subfloat[]{
	    \includegraphics[width=0.245\textwidth]{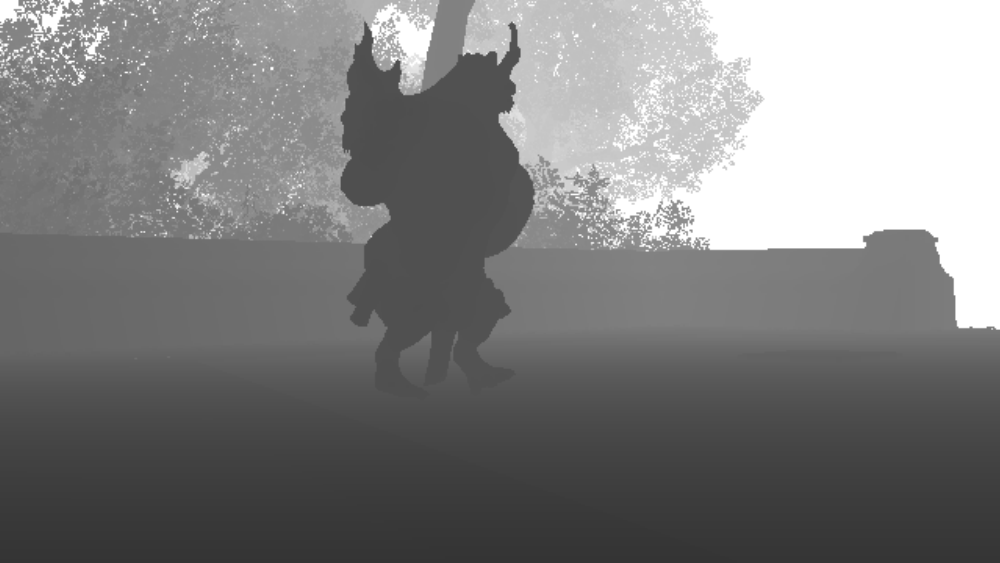}
	    \label{fig:SceneDepth}
    }
    \\[-2.5ex]
    \subfloat[]{
	    \includegraphics[width=0.25\textwidth]{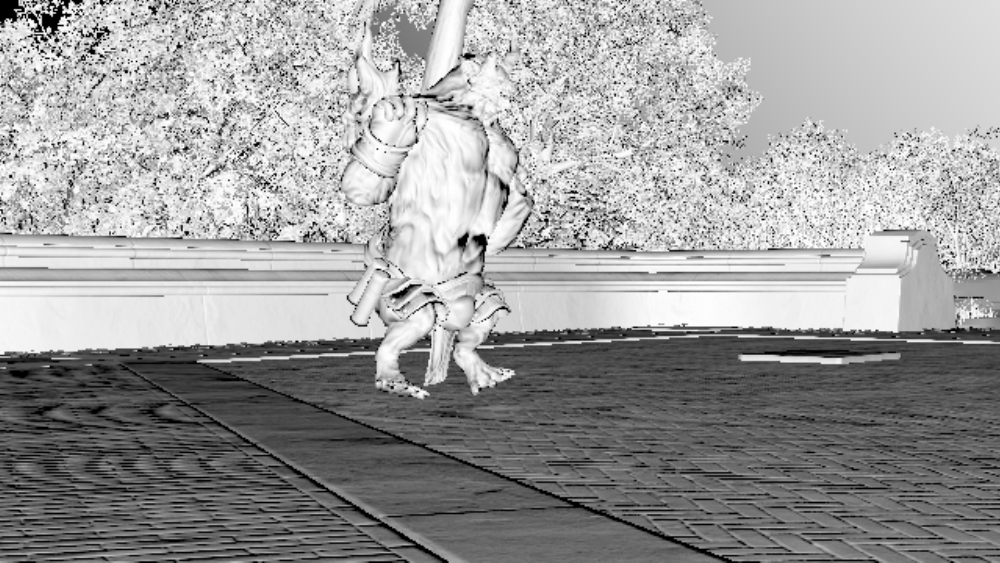}
	    \label{fig:NoV}
    }
    \subfloat[]{
	    \includegraphics[width=0.25\textwidth]{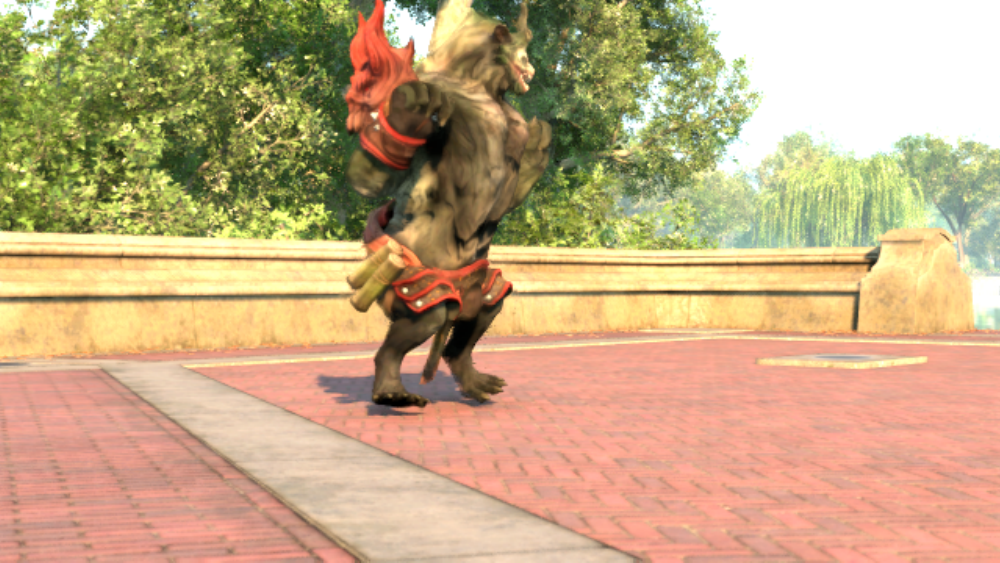}
	    \label{fig:Pretonemap}
    }
    \subfloat[]{
	    \includegraphics[width=0.25\textwidth]{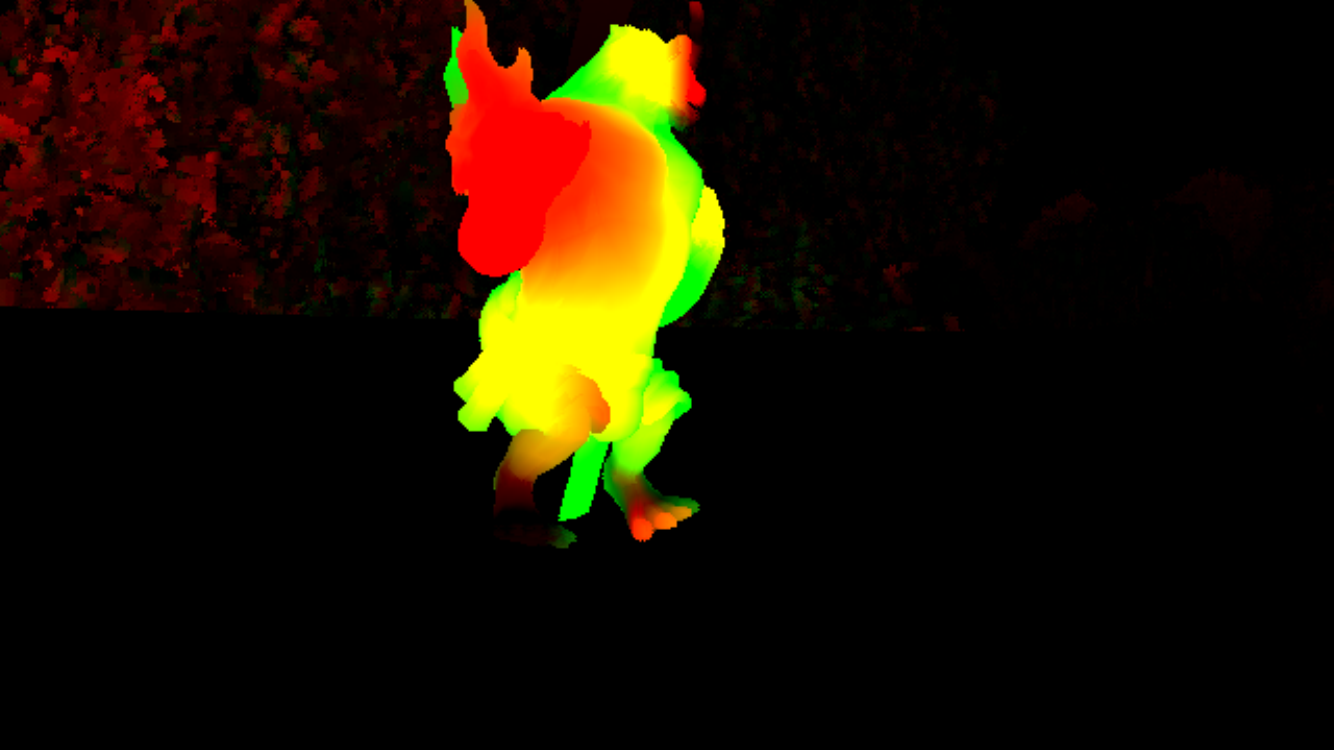}
	    \label{fig:MotionVector}
    }
    \subfloat[]{
	    \includegraphics[width=0.245\textwidth]{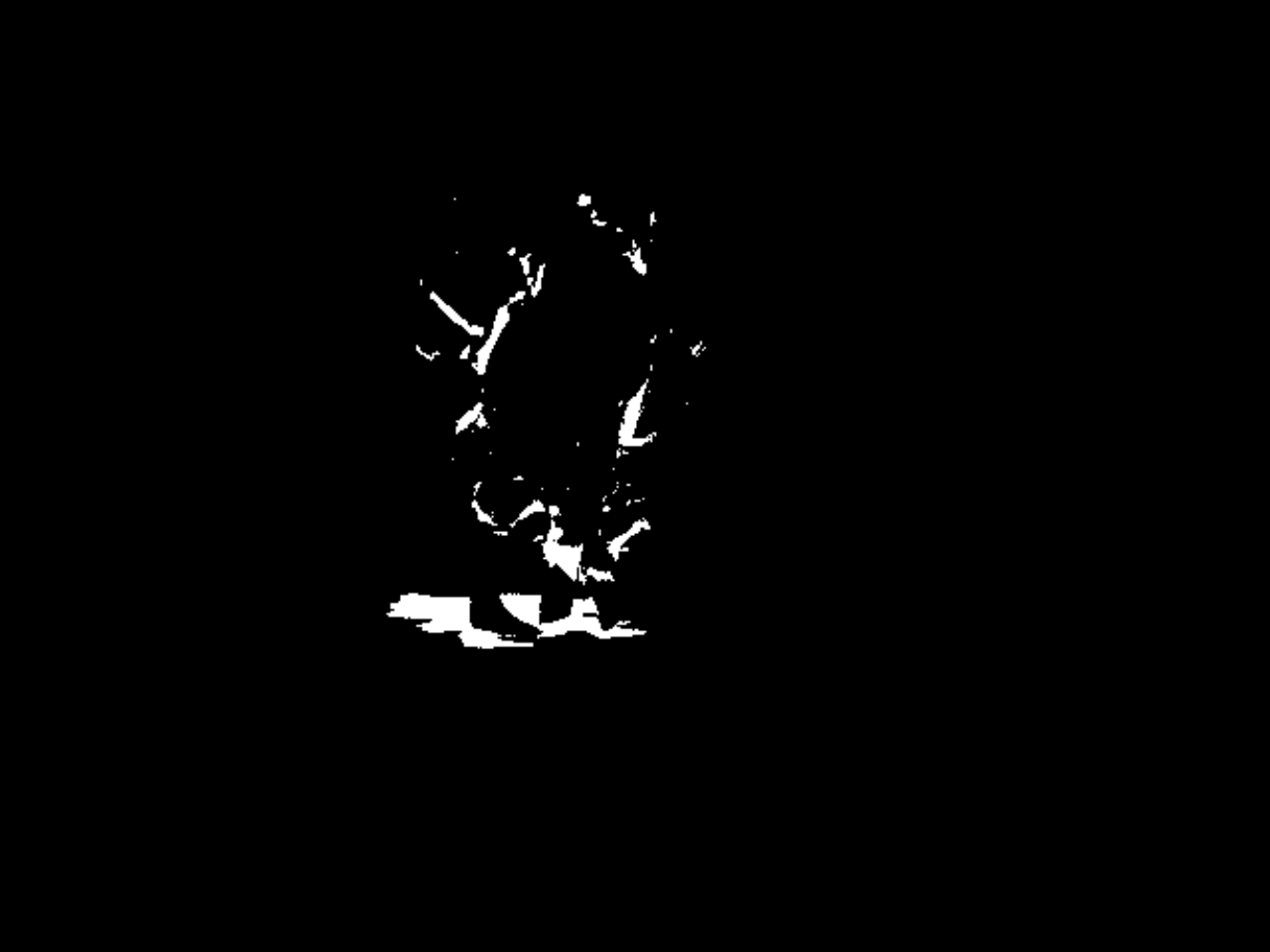}
	    \label{fig:ShadowMask}
    }
    \caption{\protect\subref{fig:stencil} Stencil; \protect\subref{fig:worldnormal} World Normal; \protect\subref{fig:worldposition} World Position; \protect\subref{fig:SceneDepth} Scene Depth; \protect\subref{fig:NoV} NoV; \protect\subref{fig:Pretonemap} PretonemapHDRColor; \protect\subref{fig:MotionVector} Motion Vector; \protect\subref{fig:ShadowMask} Shadow Mask.}
    \label{fig:buffers}  
\end{figure*}

For every frame, we extract the following G-buffers from the Unreal engine: scene depth, world normal, world position,
custom stencil, pretonemap, NoV (normal over view vector) and motion vector. A visualization of all these buffers is shown in Fig.~\ref{fig:buffers}.

\subsection{Mask Generation for Segmentation: Create Bounding Boxes}
\label{subsec:segmentation}

Based on \textbf{Insights}~\ref{box:fov_seg} and \ref{box:lat}, we propose to divide the frame into three patches and
apply different extrapolation algorithms on each patch to generate a high-quality extrapolated frame quickly.
First, we divide the frame into the foreground and background regions, followed by further dividing the background into
near-BE and far-BE
regions to adjust for the ``near-object effect''.

Foreground detection is a complex task in video processing, but with access to G-buffers, we can efficiently separate
the foreground from the background. By extracting a stencil buffer from the Unreal Engine, which stores masks for
dynamic objects in the scene, we can identify and isolate these objects~\cite{extranet}. To further classify the
background into near and far regions, we utilize the motion vector information. This classification is crucial for
distinguishing regions close to the moving object from the static background, allowing us to identify areas most
affected by foreground movements and potentially containing invalid pixels due to disocclusion. However, several
challenges arise in separating the near and far background regions. First, determining the bounding box that encompasses
the
moving object is essential for demarcating the near-BE region. Additionally, given the large variety of dynamic
objects that scenes may contain, it is vital to ensure that the background separation method is adaptable to different
scenarios.

\begin{figure}[!htbp]
	\centering
	\includegraphics[width=0.7\columnwidth]{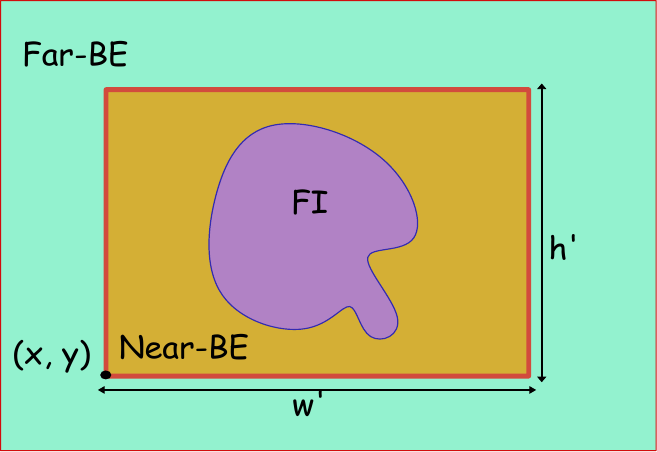}
	\caption{Identified regions in a frame. Foreground interactions are represented by FI, while Near-BE and Far-BE refer to near and far background environments, respectively. They are
    separated by a rectangular bounding box of height $h'$ and width $w'$.}
	\label{fig:regions}
\end{figure}

As mentioned in Section~\ref{subsec:math_prop}, we can segment the near and far-BE regions based on the 
temporal variation. However, the challenge lies in the real-time computation and segmentation of these regions. Hence, we
propose a simpler heuristic-based approach to perform the segmentation. Based on \textbf{Insight~\ref{box:inter_frame}},
we focus on the movement of dynamic objects to identify the near-BE region. To differentiate between the near and far
background, we start by assigning a default bounding box (based on the stencil buffer's values) 
to every moving object. To ensure a broader coverage area, we
expand the rectangle's width and height by biases $w_b$ and $h_b$, respectively, as shown in Fig.~\ref{fig:regions}.
These biases are not preset but are calculated based on the dynamic object's motion. Algorithm~\ref{alg:nearfar}
outlines the complete procedure for the background classification process.

\begin{algorithm}
    \caption{Dynamic Region Expansion for Background Classification}
    \label{alg:nearfar}
    \begin{algorithmic}[1]
    \REQUIRE Initial bounding rectangle surrounding the dynamic object $\text{Rect} = (x, y, w, h)$, motion vectors $\vec{v_x}$ and $\vec{v_y}$, scaling factors $k_x$ and $k_y$
    \ENSURE Expanded bounding rectangle $\text{Rect}' = (x, y, w', h')$
    \STATE Compute average motion (horizontal and vertical):
    \[
    \overline{v_x} = \frac{1}{N} \sum_{i=1}^{N} \vec{v_x}(i), \text{ }\overline{v_y} = \frac{1}{N} \sum_{i=1}^{N} \vec{v_y}(i)
    \]
    \STATE Determine expansion biases:
    \[
    w_b = k_x \cdot \overline{v_x}, \text{ } h_b = k_y \cdot \overline{v_y}
    \]
    \STATE Expand the bounding rectangle:
    \[
    w' = w + w_b,\text{ } h' = h + h_b
    \]
    \STATE Output the expanded rectangle: $\text{Rect}' = (x, y, w', h')$
    \end{algorithmic}
\end{algorithm}

\subsection{Data Preprocessing}
\label{subsec:preprocess}
We have already discussed the process of extracting data from the Unreal Engine and creating binary masks to segment the
frame into foreground, near-BE, and far-BE regions, respectively. As we plan to use different techniques for
extrapolation in these three regions, we require three sets of input data. Therefore, it is necessary to preprocess the
data obtained from Unreal to ensure that we have the required data in the required format. 

As we have discussed earlier, to get a quick initial estimate of the final frame, 
we {\em warp} frame $F_t$. Prior to doing this, we need to demodulate (remove many graphics effects) 
the frame: previous works have
shown that inpainting networks produce better results with demodulation~\cite{extrass,jointss}. 
The demodulation formula is shown
in Equation~\ref{eq:demodu}. As mentioned in Section~\ref{sec:RelatedWork}, we opt for the G-buffer guided warping method
because of its minimal ghosting effects. Subsequently, we detect invalid pixels within the warped frame by utilizing
G-buffers (as described in ExtraNet~\cite{extranet}). Upon completion of this step, we partition the warped frame into
three distinct regions utilizing the segmentation masks (Section~\ref{subsec:segmentation}) 
to generate inputs for the next stage.

\begin{equation}
    F' = F / ( Albedo + Specular * 0.08 * (1- Metallic)) 
  \label{eq:demodu}
\end{equation}
$F$ is the rendered image and $F'$ is the image generated after demodulation.

\subsection{Extrapolation}
\label{subsec:nn}

\subsubsection{Network Architecture}
\label{subsubsec:nn_arch}

\begin{figure*}[!h]
	\centering
	\includegraphics[width=0.99\textwidth]{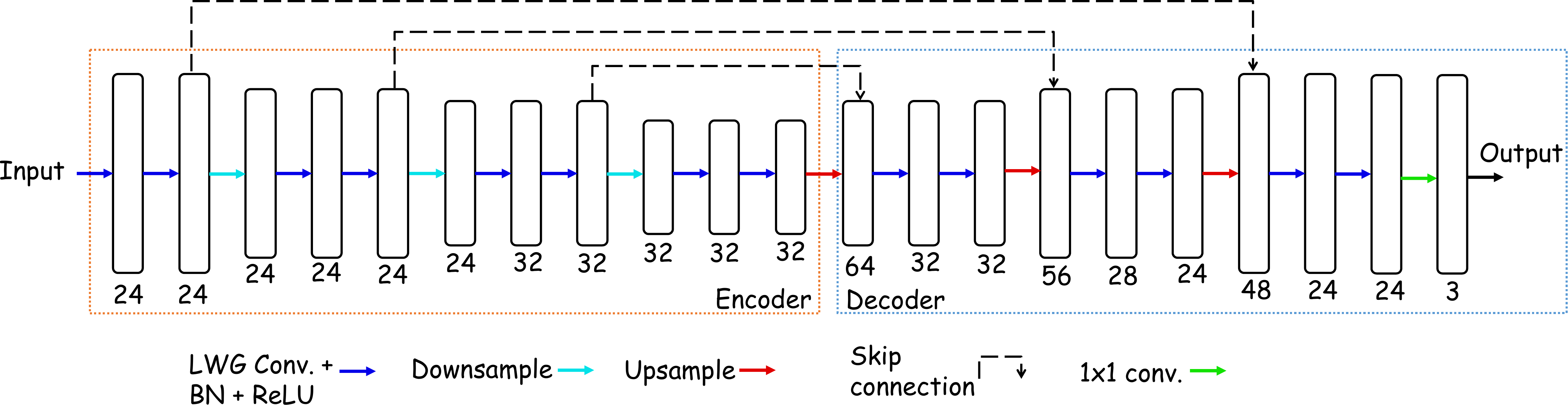}
	\caption{The neural network architecture for extrapolating the foreground region.}
	\label{fig:unet_foreground}
\end{figure*}

\begin{figure}[!h]
	\centering
	\includegraphics[width=\columnwidth]{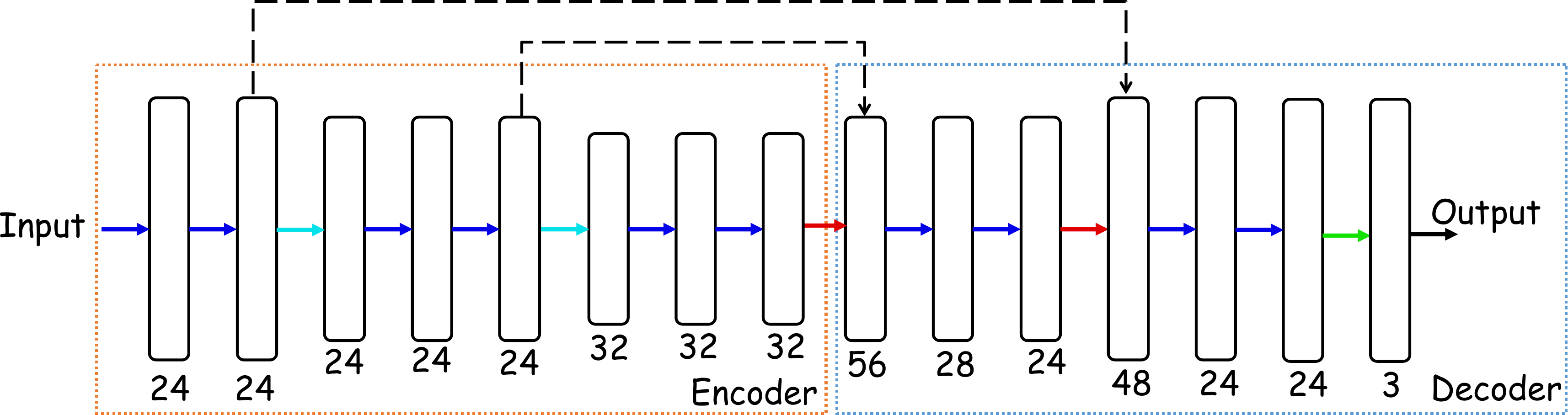}
	\caption{The neural network architecture for extrapolating the near-BE region.}
	\label{fig:unet_nearbe}
\end{figure}

We propose two bespoke neural networks for the foreground and near-BE region, respectively. 
For the far-BE region, we directly
use the warped frame. The architectures of these two networks are shown in Fig.~\ref{fig:unet_foreground} and
Fig.~\ref{fig:unet_nearbe}, respectively. Our networks are roughly similar to U-Net~\cite{ronneberger2015u} but
have important differences.
U-Net has an encoder-decoder architecture. Instead of using simple convolution, we have used
light-weight gated convolution proposed by Yi et al.~\cite{yi2020contextual} in our network which is formulated as:

\begin{equation}
  \begin{aligned}
     G (gating) =& Conv(W_g, I) \\
     F (features) =& Conv(W_f, I) \\
     O =& \sigma (G) \bigodot F 
  \end{aligned}
  \label{eq:gated}
\end{equation}

$W_g$ and $W_f$ denote two distinct learnable filters. $\bigodot$ denotes Hadamard (element-wise) multiplication, and
$\sigma$ represents the sigmoid activation function. The latter
ensures that the output gating values are in the range $[0,1]$.
This approach helps in treating different pixels differently in the network since
there are \underline{invalid pixels} in the warped
frame. Our gating mechanism diminishes the influence of invalid pixels. In our neural networks, we
utilize a comprehensive set of inputs: the warped frame, a hole mask indicating invalid pixels, and two G-Buffers
($Roughness$ and $Metallic$). Unlike previous works, we also provide as
input the LBP (local binary pattern) feature map of the
warped frame. By incorporating the LBP feature map, we leverage its robust feature extraction capabilities and its
resilience to uneven illumination. This addresses the challenge of extracting detailed features from images
with inconsistent lighting, significantly boosting the generalizability and effectiveness of our method.

\subsubsection{Loss Functions}
\label{subsubsec:loss_fns}
The {\em loss function} used for training the networks has broadly two components. The first component penalizes the pixel-wise
error between the ground truth $F$ and the predicted frame $F'$. The second component is the perceptual
loss, which was not considered in previous works~\cite{extranet,jointss}. The {\em perceptual loss} plays a crucial role in
enhancing the performance of neural networks, particularly in tasks related to image inpainting~\cite{multistep} because
it focuses on capturing high-level perceptual features, mimicking human visual perception.

\begin{equation}
    \begin{aligned}
       \mathcal{L}  =& \mathcal{L}_{pixel}  + \mathcal{L}_{perceptual}
    \end{aligned}
    \label{eq:loss}
\end{equation}

\noindent
\begin{itemize}
    \item \textbf{Pixel-wise Errors:} To calculate the error on a pixel-by-pixel basis, we use the $\mathcal{L}_1$ loss,
    which can be computed using the formula in Equation~\ref{eq:l1}. We employ the $\mathcal{L}_1$ loss in three
    different forms. First, we calculate the total $\mathcal{L}_1$ loss between the entire ground truth frame and the
    predicted frame. Second, we calculate the error between the pixels that were marked as holes or invalid pixels
    during the data preparation stage. Finally, we compute the error between the valid pixels of both frames. The total
    loss is the weighted average of these three losses (refer to Equation~\ref{eq:pixel}).

    \begin{equation} 
        \begin{aligned}
           \mathcal{L}_{\mathcal{L}_1}  =& \lVert X - Y \rVert_1 \\
           \mathcal{L}_{\mathcal{L}_1}  =& \sum_{i=1}^{H} \sum_{j=1}^{W} \lvert X(i, j) - Y(i, j) \rvert \\
        \end{aligned}
        \label{eq:l1}
    \end{equation}

    \begin{equation} 
        \begin{aligned}
           \mathcal{L}_{\mathcal{L}_1}  =& \lVert F - F' \rVert_1 \\
           \mathcal{L}_{hole}  =& \lVert (F - F')\cdot(1-m) \rVert_1 \\
           \mathcal{L}_{valid}  =& \lVert (F - F')\cdot m \rVert_1 \\
           \mathcal{L}_{pixel}  =& \lambda_{\mathcal{L}_1} \cdot\mathcal{L}_{\mathcal{L}_1} + \lambda_{hole} \cdot\mathcal{L}_{hole} + \lambda_{valid}\cdot \mathcal{L}_{valid} \\
        \end{aligned}
        \label{eq:pixel}
    \end{equation}
    
    $m$ is a binary mask used for identifying the invalid pixels. $\lambda_{\mathcal{L}_1}$, $\lambda_{hole}$, and
    $\lambda_{valid}$ are the weights assigned to each component loss function (balance out their effects). In our current implementation,
    $\lambda_{\mathcal{L}_1}$ is set to 1 and the rest of the two values are set to 0.5 each.
    
    \item \textbf{Perceptual Losses:}
    We adopt the perceptual loss~\cite{johnson2016perceptual} to guide the neural network to generate an image that is more in
    line with human perception (refer to Equation~\ref{eq:vgg}). 
     
    \begin{equation}
        \begin{aligned}
    \mathcal{L}_{VGG} =& \mathbb{E} \left[ \sum_{i} \lVert \Phi_i(F) - \Phi_i(F') \rVert_1 \right] \\
        \end{aligned}
        \label{eq:vgg}
    \end{equation}
   
    $\Phi_i$ is the activation map of the $i^{th}$ layer of the VGG-16~\cite{7486599} network pre-trained on ImageNet.
    Furthermore, we use a similar loss function called {\em  style loss}
    to maintain a degree of similarity between the predicted and
    the original image (refer to Equation~\ref{eq:style}). 
    The joint perceptual loss is shown in Equation~\ref{eq:perc}. The key idea is that we are preventing {\em hallucination},
    where the output of the model can be very different from the original image. The main aim is to fix the base image
    created by warping and not make unconstrained errors.
    
    \begin{equation}
        \begin{aligned}
    \mathcal{L}_{style} =& \mathbb{E} \left[ \sum_{i} \lVert G_j^{\Phi_i}(F) - G_j^{\Phi_i}(F') \rVert_1 \right] \\
        \end{aligned}
        \label{eq:style}
    \end{equation}
    
    $G_j^{\Phi}$ is the Gram matrix of the VGG features extracted for the perceptual loss.

    \begin{equation}
        \begin{aligned}
            \mathcal{L}_{perceptual} =& \lambda_{VGG} \cdot \mathcal{L}_{VGG} + \lambda_{style}\cdot\mathcal{L}_{style}
        \end{aligned}
        \label{eq:perc}
    \end{equation}
    
    $\lambda_{VGG}$ and $\lambda_{style}$ are the weights assigned to each loss to balance their effects. In our current implementation, $\lambda_{VGG}$ is set to 0.1 and $\lambda_{style}$ is set to 0.01.
\end{itemize}

\subsubsection{Training and Testing}
\label{subsubsec:train}

In our work, we utilize 13 benchmark scenes as discussed in Section~\ref{subsec:data}. These scenes are divided into two
groups: nine for training and four for testing. For each scene, we have created five animation sequences for training
and three sequences for testing. The number of frames in each sequence varies. The total number of frames for training
and testing for each scene are specified in Table~\ref{tab:train_test_data}. We implemented both the neural networks
using the PyTorch framework~\cite{paszke2019pytorch}. To divide the data for training and validation in an 80:20 ratio,
we utilized PyTorch's inbuilt \textit{random\_split} function.  

\begin{table}[]
    \caption{Statistics of the training and testing dataset}
    \footnotesize
    \begin{center}	
    % \resizebox{0.49\textwidth}{!}{
      \begin{tabular}{|l|c|c|c|c|}
        \hline
        %\rowcolor{gray}
        \multirow{2}{*}{\textbf{Scenes}} &{ \textbf{Training}} &  {\textbf{Training}} & \textbf{Testing} & \textbf{Testing}    \\
       &{ \textbf{Sequences}} &  {\textbf{Frames}} & \textbf{Sequences} & \textbf{Frames}    \\
        
      \hline
      \textit{PR} & 5 & 5000 & 3 &  3000 \\
      \hline 
        
        \textit{WT}  & 5 & 5000 & 3 &  3000  \\
        \hline 
        \textit{RF}  & 5 & 5000 & 3 & 3000   \\
        \hline
        \textit{CM}  & 5 & 7000 & 3 & 3000  \\
        \hline
        \textit{BR} & 5 & 5000 & 3 & 3000  \\
        \hline
        \textit{DW}  & 5 & 4000 & 3 & 3000  \\
        \hline
        \textit{TC}  & 5 & 4000 & 3 &  3000  \\
        \hline  
        \textit{TN}  & 5 & 5000 & 3 & 3000  \\
        \hline
        \textit{BK}  & 0 & 0 & 3 & 3000  \\
        \hline 
        \textit{LB} & 0 & 0 & 3 & 3000  \\
        \hline 
        \textit{TR}  & 0 & 0 & 3 & 3000    \\
        \hline
        \textit{VL}  & 0 & 0 & 3 & 3000 \\
        \hline
        \textit{SL}  & 0 & 0 & 3 & 3000 \\
        \hline
        \end{tabular}
    %    }
     \end{center} 
    \label{tab:train_test_data}
    \end{table} 

\subsubsection{Training Details}
\label{subsubsec:train_detail}
Both our neural networks are implemented and trained using the PyTorch framework~\cite{paszke2019pytorch}. We utilize
the Adam optimizer for optimization with batch sizes set to 16 and epoch sizes to 100. The network initialization
is as per the default settings in PyTorch. 

\subsection{Blending}
\label{subsec:blending}
In the final stage of the process, we bring together all three extrapolated regions: FI, near-BE, and far-BE. As
previously mentioned, we first demodulate the frame before carrying out the warping process and feed it to the neural
networks. To blend the three extrapolated regions, we use a mask-based image blending algorithm~\cite{xiong2009mask}, which is a type of weighted blending. The final blending equation is shown in Equation~\ref{eq:mask_blend}. Consequently, we need to modulate the final merged frame back using Equation~\ref{eq:modu}. Before modulation, we also need to blend the warped shadow, which will be explained in Section~\ref{subsec:shad_extra}.

\begin{equation}
    F = M_1 * F_1 + M_2 * F_2 + M_3 * F_3
  \label{eq:mask_blend}
\end{equation}
$M_1$, $M_2$ and $M_3$ are the segmentation masks (binary masks) for the foreground, near-BE and far-BE regions, respectively. $F_1$, $F_2$ and $F_3$ are the extrapolated versions of those regions.  

\begin{equation}
    F' = F * ( Albedo + Specular * 0.08 * (1- Metallic)) 
  \label{eq:modu}
\end{equation}

\subsection{Extrapolation of Shadows}
\label{subsec:shad_extra} 
%FIXME: We should have a subsection on shadows. More details on blending.

In the first stage, the data extraction stage, we extract the shadow mask for all dynamic objects using a custom blueprint class in the Unreal Engine. In the preprocessing stage, we process the shadow mask obtained from Unreal Engine. As previously discussed, we perform shadow warping to compute the extrapolated shadow. We begin by estimating the motion flow between shadows using Farneback's algorithm~\cite{farneback2003two}. In the extrapolation stage, we use the computed motion flow to warp the shadow. As discussed in Section~\ref{subsec:perf_warp}, traditional warping produces a ghosting effect in the warped frame. Therefore, we use occlusion motion vector-based warping. It is important to note that this process runs in parallel with the inpainting networks. Finally, in the blending stage, we blend the warped shadow with the blended extrapolated regions (the output of Equation~\ref{eq:mask_blend}) using an additive blending approach, which simply adds the two images on a pixel-by-pixel basis.  In the end, the resulting image is modulated using Equation~\ref{eq:modu}.

%% file: evaluation.tex
\section{Results and Analysis}
\label{sec:Evaluation}

% To showcase the performance of our proposed method, \patchex, we compare it to various state-of-the-art techniques in both frame interpolation and extrapolation. Since these existing solutions are based on deep learning, we will fine-tune their models using our dataset and then assess their performance.  Additionally, we perform ablation studies to highlight the significance of each component in our method.

To demonstrate the performance of the proposed method, \patchex, we compare it against various state-of-the-art works in
both the domains of frame interpolation and extrapolation. Since all these works are ML-based solutions, we fine-tune
their proposed neural networks for our dataset before comparing the performance. After that, we perform an ablation
study to evaluate the contribution of various individual components to the overall performance. Next, we measure the
runtime latency for each component of \patchex. For all these experiments, we use the same system configuration (refer
to Table~\ref{tab:config}).

\subsection{Performance Metrics}
\label{subsec:perf}

To measure the performance of \patchex, we use three widely used performance metrics: PSNR (Peak Signal-to-Noise Ratio),
SSIM (Structural Similarity Index), and LPIPS (Learned Perceptual Image Patch Similarity)~\cite{belhe2023discontinuity,
paliwal2023reshader}. 

PSNR, as its name suggests, is a ratio between the maximum possible power for an image and the power of the noise signal
present in the image (refer to Equation~\ref{eq:psnr}). This means that higher PSNR values signify higher quality.
However, PSNR measures the quality of images globally while ignoring local distortions. Hence, we cannot solely depend
upon PSNR to assess the performance.

Next, we use SSIM, which computes the structural similarity between two images by capturing the local patterns and
textures. It also captures the brightness and contrast information. SSIM is computed using Equation~\ref{eq:ssim}.
SSIM values range from -1 to 1, where 1 indicates perfect similarity. 

Unlike PSNR and SSIM, which simply perform pixel-wise comparison, the next performance metric we use, LPIPS,
computes the perceptual similarity using an ML model. It uses a deep neural network that extracts features from the
images and then compares the extracted features. A comparison in the feature domain aligns better with human perception. LPIPS values range from 0 to 1, and higher LPIPS values indicate that the images are more dissimilar.

\begin{equation}
    \begin{aligned}
    PSNR =& 10 \times log_{10} \left ( MAX^2/MSE \right )\\
    \end{aligned}
    \label{eq:psnr}
\end{equation}

where,
\begin{itemize}
    \item MAX is the maximum possible pixel value of the image (typically 255 for 8-bit images).
    \item MSE is the mean squared error between the original and reconstructed images.
\end{itemize}

\begin{equation}
    \begin{aligned}
    \text{SSIM}(x, y) = & \frac{{(2\mu_x\mu_y + c_1)(2\sigma_{xy} + c_2)}}{{(\mu_x^2 + \mu_y^2 + c_1)(\sigma_x^2 + \sigma_y^2 + c_2)}}
    \end{aligned}
    \label{eq:ssim}
\end{equation}
where,
\begin{itemize}
    \item $x$ and $y$ are input images. 
    \item $\mu_x$ and $\mu_y$ are the arithmetic means of $x$ and $y$, respectively.
    \item  $\sigma_x^2$ and $\sigma_y^2$ are the variances of $x$ and $y$, respectively.
    \item $\sigma_{xy}$ is the covariance of $x$ and $y$.
    \item $c_1$ and $c_2$ are small constants (to avoid division by zero).
\end{itemize}

\subsection{Performance Comparison with the Frame Extrapolation Methods}
\label{subsec:sota_extra}
In this section, we compare the performance of ExtraNet~\cite{extranet} and ExtraSS~\cite{extrass} with our proposed
method, \patchex, both qualitatively and quantitatively. As mentioned in Section~\ref{subsubsec:extrapolation}, ExtraNet
and ExtraSS are the two state-of-the-art methods that perform frame extrapolation in real-time. However, ExtraSS does
not solely extrapolate in the temporal domain; it also extrapolates in the spatial domain. Since we are dealing with
temporal supersampling, we only consider its temporal component for the purpose of comparison. 

% In this section, we provide a detailed comparison of the performance of \patchex with two other extrapolation-based methods, ExtraNet~\cite{extranet} and ExtraSS~\cite{extrass}. It is important to note that ExtraSS, as mentioned in Section~\ref{subsubsec:extrapolation}, is not a technique that extrapolates solely in the temporal domain. Instead, it provides a joint solution for extrapolation in both the temporal and spatial domains. However, for our comparison, we will be using only the temporal part of ExtraSS.

\subsubsection{Qualitative Comparisons}
\label{subsec:qual_comp_extra}

\begin{figure*}[!h]
	\centering
	\includegraphics[width=0.8\textwidth]{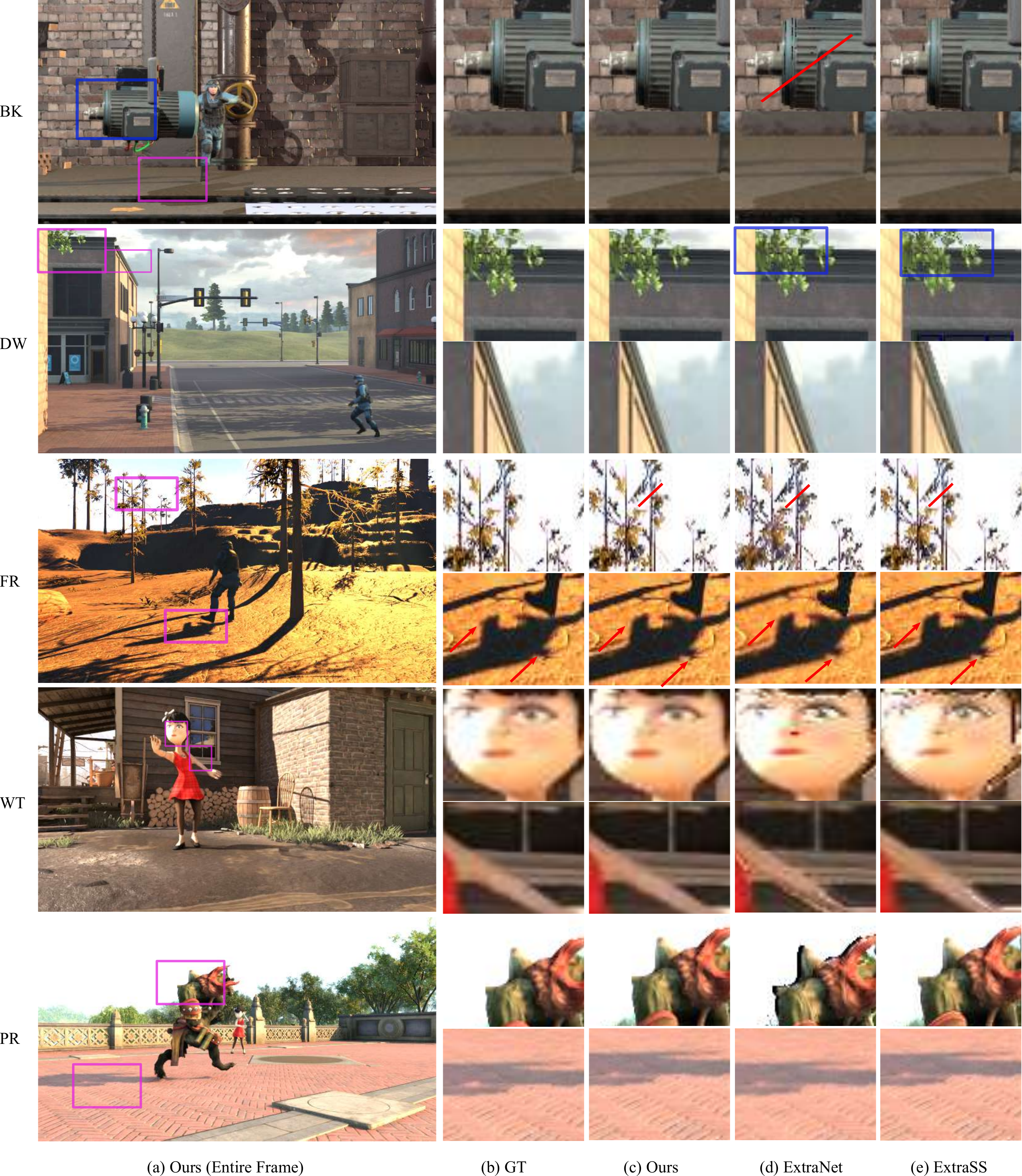}
	\caption{Visual comparisons against two frame extrapolation methods: 
  ExtraNet~\cite{extranet} and ExtraSS~\cite{extrass}}
	\label{fig:qual_comp_extra}
\end{figure*}

In this section, we compare the quality of the frame generated using various extrapolation methods. In
Fig.~\ref{fig:qual_comp_extra}, we show the extrapolated frames for five distinct benchmark scenes. Out of these five
scenes, only one scene, \textit{DW}, is captured with moving camera settings. For the rest of the scenes, the camera
is static, only the objects are moving. 

As explained in the insights in Section~\ref{box:challenges}, 
the challenges for extrapolation algorithms are to properly fill the
disoccluded regions created by the movements in the scene and to extrapolate the shadows accurately. 
Both ExtraNet and
ExtraSS fail to address these challenges effectively in many scenarios. For example, ExtraNet does not generate accurate
shadows for complex movements such as the kick in the \textit{FR} scene. It also performs poorly in capturing complex
structures such as tree leaves and facial features in scenes like \textit{BK, FR}, and \textit{WT}. On the other hand,
ExtraSS leverages G-buffer-guided warping and performs better than ExtraNet in most cases. However, it is unable to
correctly extrapolate facial features during intricate motions like those in a hip-hop dance. In the \textit{DW} scene,
where the camera is moving, and parts of the scene move out of the screen, ExtraSS handles the out-of-screen areas well,
whereas ExtraNet does not. \patchex handles all of these complex cases pretty well. To summarize, \patchex excels by not
only preserving sharp features and intricate geometries but also generating plausible shadows that closely match the
ground truth. We have uploaded a video of our results, which can be accessed using this link~\cite{results}.

% This section presents a qualitative comparison. To highlight the differences, we showcase frames from five distinct
% scenes in Fig.~\ref{fig:qual_comp_extra}, emphasizing the visual quality and effectiveness of each approach. Except
% for the \textit{DW} scene, where the camera is moving, all other scenes are static with only foreground objects in
% motion.

% While ExtraNet uses a history encoder to extrapolate shading information accurately, it struggles with generating
% shadows for complex movements, such as the kick in the \textit{FR} scene. It also performs poorly in capturing complex
% structures like tree leaves and facial features in scenes like \textit{BK, FR,} and \textit{WT}. On the other hand,
% ExtraSS leverages G-buffer-guided warping and performs better than ExtraNet in most cases. However, it is unable to
% correctly extrapolate facial features during intricate motions like those in a hip-hop dance. In the \textit{DW}
% scene, where the camera is moving and parts of the scene move out of the screen, ExtraSS handles the out-of-screen
% areas well, whereas ExtraNet does not. \patchex handles all of these complex cases pretty well. In summary, \patchex
% excels by not only preserving sharp features and intricate geometries but also generating plausible shadows that
% closely match the ground truth. We have uploaded a video of our results, which can be accessed using this
% link~\cite{results}.

\begin{table*}[!h]

  \caption{Quantitative comparison of various extrapolation methods against \patchex in terms of PSNR (dB), SSIM, and LPIPS.}
  
  \footnotesize
  \begin{center}
    
 %\resizebox{0.9\textwidth}{!}{
    \begin{tabular}{|c|ccc|ccc|ccc|}
      \hline
  %    \rowcolor{gray}
   { \multirow{2}{*}{\textbf{Scenes}}}  &\multicolumn{3}{c|}{\textbf{PSNR (dB) $\uparrow$}} &  \multicolumn{3}{c|}{\textbf{SSIM $\uparrow$}} &  \multicolumn{3}{c|}{\textbf{LPIPS $\downarrow$}}\\
     
   &   {\textbf{ExtraNet}} &   {\textbf{ExtraSS}} & {\textbf{\patchex}} &   {\textbf{ExtraNet}} &   {\textbf{ExtraSS}} & {\textbf{\patchex}} &   {\textbf{ExtraNet}} &   {\textbf{ExtraSS}} & {\textbf{\patchex}} \\
      \hline
     
    PR & 28.78  & 21.87 & {\bf 36.49}& 0.909 & 0.948 & {\bf 0.988} & 0. 111 & 0.118 & {\bf 0.006}\\
    BK &  20.21  & 24.26 & {\bf 37.38} & 0.880 &  0.906 & {\bf 0.991} & 0.190& 0.185& {\bf 0.007} \\
     
    WT &  24.01  & 23.75 &  {\bf 40.38} &  0.848 & 0.796 & {\bf 0.981} & 0.213& 0.109&{\bf 0.008}\\
     
    RF & 17.51  & 21.55 & {\bf 36.55} & 0.733 &  0.785& {\bf 0.995} & 0.342 & 0.261&  {\bf 0.005}\\
      
    CM & 33.24   & 22.97 & {\bf 30.72} & 0.867 &  0.502 & {\bf 0.765} & 0.226& 0.261& {\bf 0.117}\\
     
    BR & 24.69   & 26.55 &  {\bf 34.78} & 0.583 & 0.614 & {\bf 0.987} & 0.465& 0.384 &{\bf 0.009}\\
     
    DW & 19.63 & 25.03 &  {\bf 36.55} & 0.791 & 0.876 & {\bf 0.984} & 0.376& 0.169 & {\bf 0.024}\\
      
    TC & 22.36 & 23.78 & {\bf 31.63}  & 0.629 & 0.705 & {\bf 0.980}  & 0.362& 0.398& {\bf 0.022}\\
     
    LB & 24.49  & 27.54 &  {\bf 38.59} &  0.846 &  0.901 & {\bf 0.989} & 0.414& 0.091 &{\bf 0.011}\\
       
    TR & 17.40 & 27.86 & {\bf 34.92}  & 0.637 &  0.677 &  {\bf 0.985} & 0.445& 0.295& {\bf 0.021}\\
     
    VL & 22.19 & 21.00 & {\bf 39.23} & 0.817 & 0.818 &  {\bf 0.995} & 0.314& 0.178 & {\bf 0.023}\\
      
    TN & 17.39 & 26.91 & {\bf 37.35} & 0.754 & 0.815 &  {\bf 0.995} & 0.305& 0.205 & {\bf 0.009}\\
      
    SL & 16.47 & 28.12 & {\bf 42.19} &  0.467 & 0.510 & {\bf 0.998} & 0.496& 0.549& {\bf 0.054}\\
    \hline
    \end{tabular}
   %}
  \end{center} 
  \label{tab:sota_extra}
  \end{table*}

\subsubsection{Quantitative Comparisons}
\label{subsec:quant_comp_extra}
In this section, we perform a quantitative comparison among  \patchex, ExtraNet, and ExtraSS using three performance
metrics: PSNR, SSIM, and LPIPS. We show the final results in Table~\ref{tab:sota_extra} and make the following
observations:

\circled{1} Our method performs better than ExtraNet and ExtraSS across all benchmarks for all the performance metrics.

\circled{2} There is 65.29\%, 32.76\%, and 92.66\% increase in PSNR, SSIM, and LPIPS in \patchex, respectively, as compared to ExtraNet.

\circled{3} Compared to ExtraSS, our method achieves an improvement of 48.46\%, 31.53\%, and 90.24\% in PSNR, SSIM, and LPIPS, respectively.

% In addition to the qualitative analysis, we also perform a quantitative comparison of the frame extrapolation methods. Table~\ref{tab:sota_extra} presents the quantitative evaluation in terms of PSNR, SSIM, and LPIPS. From these results, we make the following observations:

% \circled{1} \patchex significantly outperforms previous frame extrapolation methods across all metrics.

% \circled{2} \patchex shows a substantial improvement over ExtraNet, achieving an improvement of 65.29\% in PSNR, 32.76\% in SSIM, and 92.66\% in LPIPS on average.

% \circled{3} \patchex demonstrates a significant improvement over ExtraSS, with an improvement of 48.46\%, 31.53\%, and 90.24\% on average in PSNR, SSIM, and LPIPS, respectively.

\subsection{Performance Comparison with Frame Interpolation Methods}
\label{subsec:sota_inter}

In this section, we compare the performance of \patchex with two state-of-the-art interpolation-based methods. The
interpolation-based methods are Softmax Splatting~\cite{softmax} and EMA-VFI~\cite{EMA}.  All these methods are
DNN-based techniques. Softmax splatting uses forward warping; it uses forward and backward motion flow (reprojection).
However, in this approach, multiple pixels may map to the same target location in frame $F_t$. Softmax splatting uses a
modified softmax layer, which takes the frame's depth data to resolve this ambiguity. EMA-VFI uses a transformer network
to perform frame interpolation. 

\subsubsection{Qualitative Comparisons}
\label{subsec:qual_comp_inter}
\begin{figure*}[!h]
	\centering
	\includegraphics[width=0.8\textwidth]{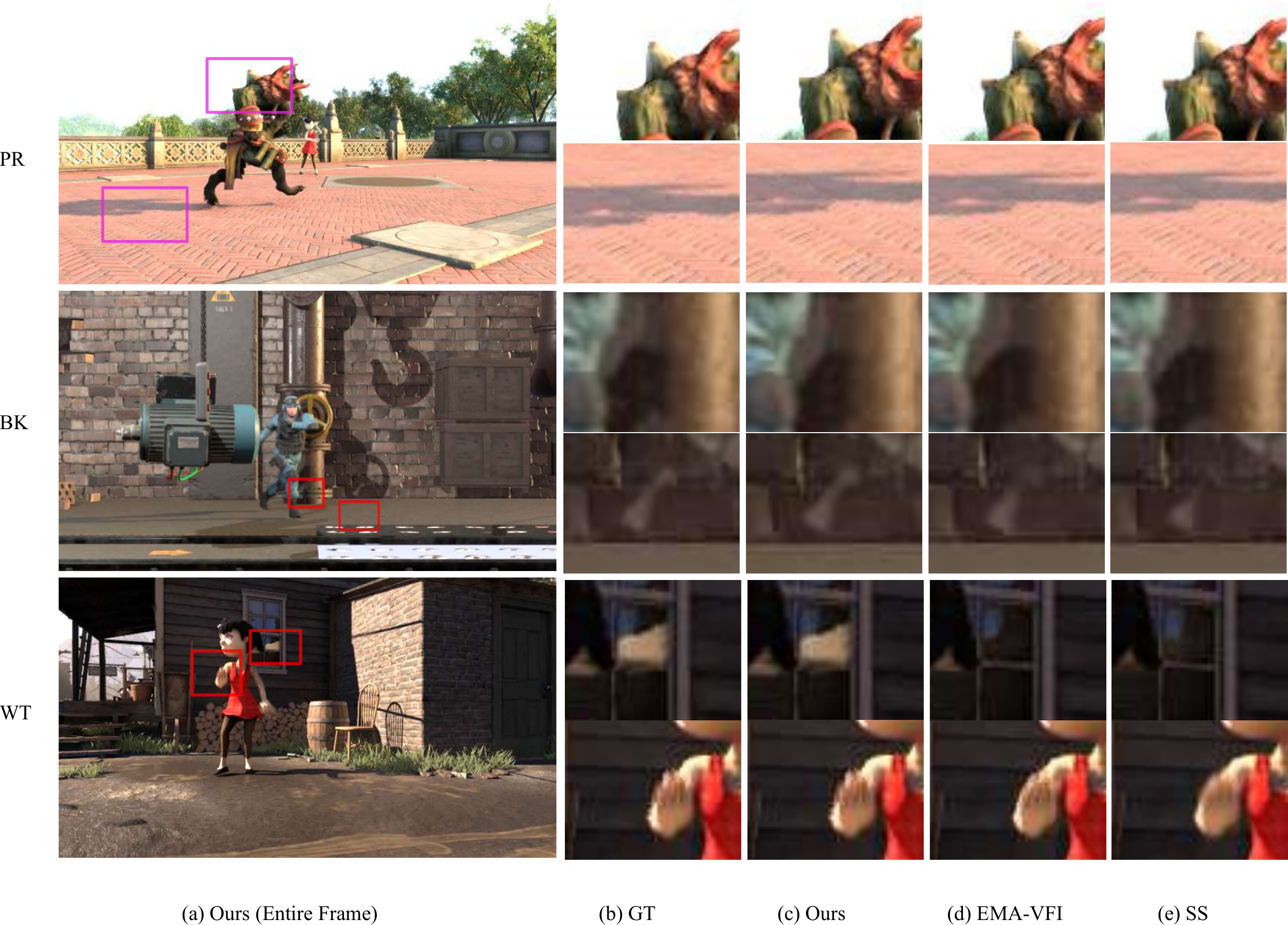}
	\caption{Visual comparisons against two frame interpolation methods: EMA-VFI~\cite{EMA} and Softmax-splatting (SS)~\cite{softmax}}
	\label{fig:qual_comp_inter}
\end{figure*}

In this section, we conduct a qualitative comparison. It is important to note that while all interpolation methods
necessitate pre-rendered future frames, our extrapolation-based method exclusively relies on historical frames that have
already been rendered. Consequently, interpolation methods generally exhibit superior performance compared to our
method; however, there are instances where they exhibit shortcomings. To illustrate such cases, we present frames from
three distinct scenes in Fig.~\ref{fig:qual_comp_inter}, emphasizing the visual quality and effectiveness of each
approach.

In the \textit{PR} scene, both interpolation-based techniques produce shadows that closely resemble the ground truth.
However, in the \textit{BK} scene, the sharp definition of the shadow structure is compromised for both these methods. This
can be attributed to the higher glossiness factor present in the \textit{BK} scene as compared to others. Notably, our
method excels in this scenario due to its utilization of G-Buffer information, which incorporates glossiness data.
Another notable artifact in interpolation methods is the potential for blurriness during complex movements, as observed
in the \textit{WT} scene. We have uploaded a video of our results, which can be accessed using this link~\cite{results}.

\subsubsection{Quantitative Comparisons}
\label{subsec:quant_comp_inter}
In addition to the qualitative analysis, we also perform a quantitative comparison of the frame interpolation methods.
Table~\ref{tab:sota_inter} presents the quantitative evaluation in terms of the PSNR, SSIM, and LPIPS
metrics. From these results,
we make the following observations:

\circled{1} Quantitatively, interpolation-based methods perform better than \patchex; however, the difference is not
significant. On average, our method shows only a 12.35\% decrease in PSNR compared to EMA-VFI and an 11.06\% decrease
compared to Softmax-splatting.

\circled{2} For SSIM, there are several instances where our method outperforms the others. This is because interpolation
can cause blurriness, which significantly impacts structural details, leading to a lower SSIM value.

\circled{3} Even for LPIPS, our method performs reasonably well, achieving an average value of 0.024 (lower the better).

\begin{table*}[!h]
\caption{Quantitative comparison of various interpolation methods against \patchex in terms of PSNR (dB), SSIM, and LPIPS.}
\footnotesize
\centering
% \resizebox{0.9\textwidth}{!}{
\begin{tabular}{|c|ccc|ccc|ccc|}
\hline

 { \multirow{2}{*}{\textbf{Scenes}}}  &\multicolumn{3}{c|}{\textbf{PSNR (dB) $\uparrow$} } &  \multicolumn{3}{c|}{\textbf{SSIM $\uparrow$}} &  \multicolumn{3}{c|}{\textbf{LPIPS $\downarrow$}}\\
   
 &   {\textbf{EMA-VFI}} &   {\textbf{SS}} & {\textbf{\patchex}} &   {\textbf{EMA-VFI}} &   {\textbf{SS}} & {\textbf{\patchex}} &   {\textbf{EMA-VFI}} &   {\textbf{SS}} & {\textbf{\patchex}} \\
    \hline
   
  PR & {\bf 40.08}  &  39.50 & 36.49 &  {\bf 0.994} & {\bf 0.994} & 0.988 & \textbf{0.002} & \textbf{0.002} & 0.006\\
   
  BK & {\bf 43.42}  & 42.82 &  37.38 & {\bf 0.996} &  {\bf 0.996} & 0.991 & 0.002 & \textbf{0.001} & 0.007\\
   
  WT & {\bf 44.87}  & 44.31 &  40.38 &  {\bf 0.996} &  {\bf 0.996} & 0.981 & 0.004 & \textbf{0.001} &  0.008\\
   
  RF & {\bf 38.49}  & 37.83 & 36.55 & 0.992 &  0.990 & \textbf{0.995} & 0.003 & \textbf{0.001} & 0.005\\
    
  CM & {\bf 43.96}   & 43.62 &  30.72  &  {\bf 0.985} &  0.982 & 0.765 & 0.006 & \textbf{0.003} & 0.017\\
   
  BR & {\bf 36.48}   & 36.27 &  34.78 &  {\bf 0.988} & {\bf 0.988} & 0.987 & 0.007 & \textbf{0.005} & 0.009\\
   
  DW & {\bf 44.04} & 43.37 & 36.55 & {\bf 0.995} & {\bf 0.995} & 0.984 & 0.001 & \textbf{0.001} & 0.024\\
    
  TC & {\bf 36.27}  & 35.79 &  31.63  & {\bf 0.991} &  {\bf 0.991} & 0.980 & 0.005 & \textbf{0.003} & 0.022\\
   
  LB &  {\bf 45.77}  & 44.91 & 38.59  & 0.996 &  {\bf 0.997} & 0.989& \textbf{0.001} & \textbf{0.001} & 0.011\\
     
  TR &  {\bf 37.07} & 36.69 &  34.92 & {\bf 0.991} &  0.990 & 0.985 & \textbf{0.004} & \textbf{0.004} & 0.021\\
   
  VL & {\bf 44.24} &  43.26 &  39.23  &  {\bf 0.997} &  {\bf 0.997} & 0.995 & \textbf{0.001} & \textbf{0.001} & 0.023\\
    
  TN & {\bf 41.48} & 40.92 & 37.35 & {\bf 0.996} &  {\bf 0.996} & 0.995 & 0.002 & \textbf{0.001} & 0.009\\
   
  SL & {\bf 47.64} & 46.66 & 42.19 &  0.994 &  0.996 & \textbf{0.998} & \textbf{0.001} & \textbf{0.001} & 0.054\\
  \hline
  \end{tabular}
 %}

\label{tab:sota_inter}

\end{table*}

\subsection{Performance Analysis for High-Resolution Frames}
\label{subsec:var_res}
In this area, the standard practice is to perform temporal supersampling at a resolution of 360p and
then use spatial supersampling to increase the resolution. All the prior work in this area~\cite{} have done the same.
Akin to our paper, they assume that the spatial supersampling technique is orthogonal. 

Nevertheless, for
gaining valuable insights into the efficiency of our algorithm let us evaluate its effectiveness when we 
{\em directly work} with frames at a full-HD resolution (1080p). This is a {\em though experiment}.
We maintain the same experimental setup and use the same
evaluation metrics.

\begin{figure}[!htbp]
	\centering
	\includegraphics[width=0.99\columnwidth]{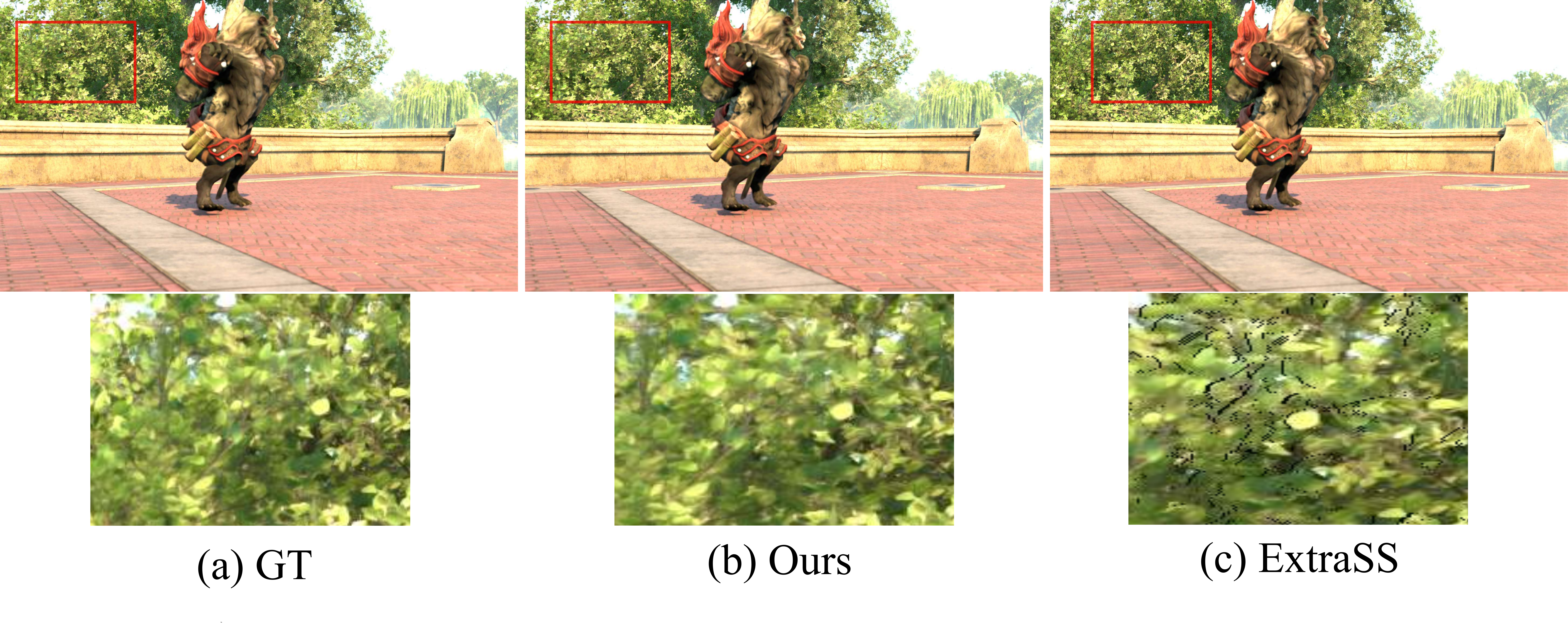}
	\caption{Performance on high-resolution frames}
	\label{fig:qual_highres}
\end{figure}

Figure~\ref{fig:qual_highres} shows that the extrapolated frames in this setting closely match the ground truth.
Table~\ref{tab:quan_var_res} presents the average PSNR, SSIM, and LPIPS values across benchmarks for various frame
resolutions. In the interest of saving space, we are not showing all the results. However, a comparison with {\em ExtraSS} that has a built-in spatial supersampler must be done. Representative results are shown in Table~\ref{tab:quan_var_res}.
%FIXME: Fill this part. 

\begin{table}[!h]

  \caption{Performance of \patchex in terms of average PSNR (dB), SSIM, and LPIPS at various resolution levels.}
  
  \footnotesize
  \begin{center}
    
 \resizebox{0.99\columnwidth}{!}{
  \begin{tabular}{|l|c|c|c|c|c|c|}
  \hline
  %    \rowcolor{gray}
   {\textbf{Scene}}  & \multicolumn{3}{c|}{\textbf{\patchex}} & \multicolumn{3}{c|}{\textbf{ExtraSS}}\\
   {\textbf{res.}}  &{\textbf{PSNR$\uparrow$} } &  {\textbf{SSIM$\uparrow$}} &  {\textbf{LPIPS$\downarrow$}} &{\textbf{PSNR$\uparrow$} } &  {\textbf{SSIM$\uparrow$}} &  {\textbf{LPIPS$\downarrow$}}\\
     \hline
     
    360p &  36.67  & 0.971  & 0.024 & 24.70 & 0.758 & 0.246 \\
     
    480p &  35.90 & 0.967  &  0.035 & 22.98 & 0.769 & 0.258 \\
     
    720p &  36.23 & 0.963 & 0.019 & 21.88 & 0.747 & 0.247 \\
     
    1080p & 36.25 & 0.969 & 0.026 & 20.77 & 0.718 & 0.218 \\
    \hline
    \end{tabular}
   }
   \end{center}
  \label{tab:quan_var_res}
\end{table}

\subsection{Ablation Study}
\label{subsec:ablation}
\begin{table*}[h]

  \caption{Quantitative comparison of various variants of \patchex in terms of PSNR (dB), SSIM, and LPIPS. \textit{w/o FS} refers to without foveated segmentation. \textit{w/o SP} refers to the case where shadow and the inpainting tasks are not handled separately. \textit{w/o $\mathcal{L}_{p}$} refers to the case where perceptual loss is not taken into account.}
  
  \footnotesize
  \begin{center}
    
 %\resizebox{0.9\textwidth}{!}{
    \begin{tabular}{|c|cccc|cccc|}
      \hline
  %    \rowcolor{gray}
   { \multirow{2}{*}{\textbf{Scenes}}}  &\multicolumn{4}{c|}{\textbf{PSNR (dB) $\uparrow$}} &  \multicolumn{4}{c|}{\textbf{SSIM $\uparrow$}} 
  %  &  \multicolumn{4}{c|}{\textbf{LPIPS $\downarrow$}}
  \\
     
   &   {\textbf{w/o FS}} &   {\textbf{w/o SP}}&   {\textbf{w/o $\mathcal{L}_{p}$}} & {\textbf{\patchex}} & {\textbf{w/o SP}} &   {\textbf{w/o FS}}&   {\textbf{w/o $\mathcal{L}_{p}$}} & {\textbf{\patchex}} 
  %  & {\textbf{w/o SP}} &   {\textbf{w/o FS}}&   {\textbf{w/o $\mathcal{L}_{p}$}} & {\textbf{\patchex}} 
  \\ \hline
     
    PR & 34.32 & 35.40 & 35.50  & 36.49 &   0.985 & 0.985&  0.985 & 0.988  \\
    BK & 33.91& 36.64 & 36.61 & 37.38 &   0.990& 0.991&   0.992 & 0.991     \\
     
    WT & 38.70& 31.55 & 39.71 & 40.38 &    0.994& 0.995& 0.984 & 0.981 \\
     
    RF & 34.43 & 35.44 & 35.43 & 36.55 &   0.991&0.993 & 0.993 & 0.995 \\
      
    CM &30.59 & 30.52 & 30.54 & 30.72 &     0.761 & 0.759 & 0.760 & 0.765  \\
     
    BR &  31.42 & 33.10 &  32.44 & 34.78 &    0.979& 0.981 & 0.979 & 0.987 \\
     
    DW &  35.04 & 35.83 & 35.82 & 36.55&    0.981 & 0.982 & 0.982 & 0.984 \\
      
    TC & 28.54 & 30.28 &  29.91 & 31.63 &    0.974& 0.979 & 0.977 & 0.980 \\
     
    LB & 34.26 & 36.75 & 37.01 & 38.59 &    0.985  & 0.988 & 0.988 & 0.989  \\
       
    TR & 31.89 & 32.50 & 32.61 & 34.92 &    0.980 & 0.981 & 0.981 & 0.985 \\
     
    VL &  36.58 & 37.55 & 38.98 & 39.23 &     0.994 & 0.987 & 0.996 & 0.995 \\
      
    TN &  33.59 & 35.92 & 36.05 & 37.35 &     0.989 & 0.991 & 0.992 & 0.995 \\
     
    SL & 40.72  & 40.12 & 41.23 & 42.19 &     0.998 &  0.994 & 0.997 & 0.998 \\
    \hline
    \end{tabular}
   %}
  \end{center} 
  \label{tab:ablation}
\end{table*}

To thoroughly understand the impact of different components in our method, we conducted an ablation study focusing on
three key elements: perceptual loss, foveated segmentation and the separate handling of shadows and inpainting. This
analysis helps to isolate the contribution of each component in the overall performance of our frame extrapolation
technique.  We evaluate three distinct variants of \patchex: one where shadow and inpainting tasks are not partitioned,
another without foveated segmentation and a third without perceptual loss. The quantitative comparison of these
variants along with the original method is shown in Table~\ref{tab:ablation}.

We make the following observations from the results:

\noindent \circled{1} These results highlight the significant impact of foveated segmentation on improving image quality. Without
foveated segmentation, the average PSNR decreases by almost 2.5 dB.

\noindent \circled{2} Similarly, we see the impact of shadow and image partitioning; there is an improvement of  1.9 dB in
PSNR. 

\noindent \circled{3} Likewise, we observe the significant impact of the perceptual loss that we included in the training of neural
networks, resulting in an improvement of 1.1 dB in PSNR. These findings remain consistent for SSIM as well. This
underscores the effectiveness of incorporating perceptual loss in improving both pixel-level fidelity and structural
similarity in the reconstructed frames.

% \subsubsection{Validation of Perceptual Loss}
% \label{subsubsec:abl_loss}
% Perceptual loss plays a crucial role in ensuring that the generated frames maintain high visual fidelity by focusing on the perceptual differences rather than just pixel-wise differences. To assess its impact, we trained our network both with and without the perceptual loss component.
 
% \subsubsection{Validation of Foveated Segmentation}
% \label{subsubsec:abl_segment}
% Foveated segmentation leverages the human visual system's focus on the foreground, optimizing the processing resources for the most perceptually important regions of the frame.

% \subsubsection{Validation of Shadow Handling}
% \label{subsubsec:abl_shadow}
% Shadows and inpainting are critical for maintaining the realism of generated frames, especially in dynamic scenes. By specifically addressing shadows and inpainting separately, the method produced more natural and coherent frames. Shadows appeared more realistic, and inpainting seamlessly filled in missing areas without noticeable artifacts.

\subsection{Latency of \patchex}
\label{subsec:latency}

As mentioned in Section~\ref{sec:methodology}, we divide \patchex into five major steps. The runtime latency of these
steps is shown in Table~\ref{tab:runtime_diffRes}. We make the following observations from the table:

\noindent \circled{1} Due to the partitioning of the frame and parallelization of the extrapolation processes, the inference latency
is impressively low. Specifically, for 360p resolution, the latency is merely 0.67 ms. Even for higher resolutions such
as 1080p, the latency remains remarkably low, averaging only about 2 ms. \\ 

\noindent \circled{2} For other components, such as
warping and preprocessing, the latency tends to increase as the frame resolution increases.

\begin{table}[!h]
  \caption{Runtime (ms) breakdown of the proposed method at various resolution levels}
  \footnotesize
  \begin{center}
     \resizebox{0.99\columnwidth}{!}{
      \begin{tabular}{|l|l|l|l|l|l|}
      \hline
     { \multirow{2}{*}{\textbf{Scenes}}}  &\multicolumn{5}{c|}{\textbf{Step}} \\
     &   {\textbf{G-Buffer}} &   {\textbf{Warping}} & {\textbf{Preprocessing}}  & {\textbf{Inference}} & {\textbf{Blending}} \\
      \hline
       
      360p & 0.17  & 0.75  & 1.44 & 0.67 & 0.01\\
       
      480p &  0.36 &  0.89 & 1.89   & 1.28 & 0.01\\
       
      720p & 1.01 & 1.58 &  2.19 & 1.98 & 0.02\\
       
      1080p &  1.21 & 2.05 & 3.12 & 2.14 & 0.03\\
      \hline
      \end{tabular}
      }
  \end{center}  
  \label{tab:runtime_diffRes}
  \end{table}

%% file: conclusion.tex
\section{Conclusion}
\label{sec:Conclusion}
With high-frequency displays becoming increasingly popular, generating frames for real-time applications at higher rates
with superior quality is necessary. Since applications are very demanding in terms of processing power, even most GPUs
cannot provide a consistently high frame rate at an HD/4K resolution. This work illustrates one such method, \patchex,
of supersampling in the temporal domain that strives to provide the quality of interpolation with the latency
of extrapolation. We propose a novel method of 
partitioning the frame into patches and parallelizing the extrapolation tasks of these
patches to reduce the latency of the temporal supersampling task. Furthermore, to improve the quality of the final
extrapolated frame, we propose using different types of extrapolation algorithms using our bespoke neural
networks for each patch. We also recognize the importance of computing shadows correctly, extrapolating them
and blending them into
the final output. We achieved an improvement of 48.46\%  in quality (PSNR) and 2$\times$ better
latency as compared to the nearest competing work.